%% file: evolutiongraph.tex
\documentclass[sigconf]{acmart}

\copyrightyear{2021}
\acmYear{2021}
\setcopyright{acmcopyright}
\acmConference[WSDM '21] {Proceedings of the Fourteenth ACM International Conference on Web Search and Data Mining}{March 8--12, 2021}{Virtual Event, Israel}
\acmBooktitle{Proceedings of the Fourteenth ACM International Conference on Web Search and Data Mining (WSDM '21), March 8--12, 2021, Virtual Event, Israel}
\acmPrice{15.00}
\acmDOI{10.1145/XXXXXX.XXXXXX}
\acmISBN{978-1-4503-8297-7/21/03}

\usepackage{booktabs} 
\usepackage{verbatim}
\usepackage{amsfonts}
\usepackage{algorithm, algorithmicx, algpseudocode}
\usepackage{amsmath}
\usepackage{array}
\usepackage{subfigure}
\usepackage{balance}
\usepackage{multicol}
\usepackage{multirow}
\usepackage{makecell}
\usepackage{color}
\usepackage{epsfig}
\usepackage{natbib}
\usepackage{xspace}
\usepackage{bm}
\usepackage{enumitem}
\usepackage{diagbox}
\usepackage{url}
\usepackage{hyperref}
\usepackage{changepage}
\newcommand{\vpara}[1]{\smallskip\noindent\textbf{#1 }}
\newcommand{\equationref}[1]{Eq~\ref{#1}}
\newcommand{\secref}[1]{Section~\ref{#1}} 
\newcommand{\alref}[1]{Algorithm~\ref{#1}} 
\newcommand{\figref}[1]{Figure~\ref{#1}} 
\newcommand{\tableref}[1]{Table~\ref{#1}} 

\newcommand{\methodname}{Evolutionary State Graph Network\xspace}
\newcommand{\graphname}{evolutionary state graph\xspace}
\newcommand{\graphnames}{evolutionary state graphs\xspace}
\newcommand{\methodshort}{EvoNet\xspace}

\newcommand{\kagglestocks}{DJIA 30 Stock Time Series\xspace}
\newcommand{\kagglestocksshort}{DJIA30\xspace}
\newcommand{\kaggletraffic}{Web Traffic Time Series Forecasting\xspace}
\newcommand{\kaggletrafficshort}{WebTraffic\xspace}
\newcommand{\telecomdataset}{Information Networks Supervision\xspace}
\newcommand{\telecomshort}{NetFlow\xspace}
\newcommand{\clockdataset}{Watt-hour Meter Clock Error\xspace}
\newcommand{\clockshort}{ClockErr\xspace}
\newcommand{\aliyundataset}{Abnormal Server Response\xspace}
\newcommand{\aliyunshort}{AbServe\xspace}

\begin{document}
	
\title{Time-Series Event Prediction with Evolutionary State Graph} 
\thanks{$^1$This work was done when the first author was studying in Zhejiang University. ~ Corresponding author: Yang Yang, yangya@zju.edu.cn \\
$^2$The code and data are publicly released at \url{https://github.com/zjunet/EvoNet}}

\author{Wenjie Hu$^{\dagger\ast}$, Yang Yang$^{\dagger}$, Ziqiang Cheng$^{\dagger}$, Carl Yang$^{\ddagger}$, Xiang Ren$^{\S}$}

\affiliation{
	\institution{$^{\dagger}$College of Computer Science and Technology, Zhejiang University, Hangzhou, China}
	\institution{$^{\ast}$Alibaba Cloud, Alibaba Group, Hangzhou, China}
	\institution{$^{\ddagger}$Emory University, Atlanta, GA, USA}
	\institution{$^{\S}$University of Southern California, Los Angeles, CA, USA}
	\institution{$^{\ast}$dulin.hwj@alibaba-inc.com, $^{\dagger}$\{aston2une, yangya, petecheng\}@zju.edu.cn, $^{\ddagger}$j.carlyang@emory.edu, $^{\S}$xiangren@usc.edu}
}
\email{}

\renewcommand{\shortauthors}{Hu et al.}

%
%


\input{abstract.tex}

\maketitle
\input{intro.tex}
\input{setup.tex}
\input{model.tex}
\input{exp.tex}
\input{related.tex}

\input{conclusion.tex}

\small
\vpara{Acknowledgments.}
Yang Yang's work is supported by NSFC (61702447), the National Key Research and Development Project of China (No. 2018AAA\\0101900), the Fundamental Research Funds for the Central Universities, and research funding from the State Grid Corporation of China. 
Xiang Ren’s work is supported by the DARPA MCS program under Contract No. N660011924033 with the United States Office Of Naval Research and NSF SMA 18-29268.
Wenjie Hu's work is supported by Alibaba Group.
\normalsize

\balance
{
	\bibliographystyle{ACM-Reference-Format}
	\bibliography{reference}
}

\pagebreak
\appendix
\input{appendix}

\end{document}

%% file: abstract.tex

\begin{abstract}
	The accurate and interpretable prediction of future events in time-series data often requires the capturing of representative patterns (or referred to as \textit{states}) underpinning the observed data.
	To this end, most existing studies focus on the representation and recognition of states, 
	but ignore the \textit{changing transitional relations} among them.
	In this paper, we present \textit{\graphname}, a dynamic graph structure designed to systematically represent the evolving relations (edges) among states (nodes) along time.
	We conduct analysis on the dynamic graphs constructed from the time-series data and show that changes on the graph structures (e.g., edges connecting certain state nodes) can inform the occurrences of events (i.e., time-series fluctuation).
	Inspired by this, we propose a novel graph neural network model, \textit{\methodname} (\textit{\methodshort}), to encode the \graphname for accurate and interpretable time-series event prediction.
	Specifically, \textit{\methodname} models both the node-level (state-to-state) and graph-level (segment-to-segment)  propagation, and captures the node-graph (state-to-segment) interactions over time. 
	Experimental results based on five real-world datasets show that our approach not only achieves clear improvements compared with 11 baselines, but also provides more insights towards explaining the results of event predictions.
	
\end{abstract}

%% file: intro.tex

\section{Introduction}
\label{sec:intro}
The prediction of future events (e.g., anomalies) in time-series data has been an important task for temporal data mining \citep{du2016recurrent, ning2016modeling, liu2019characterizing, ailliot2012markov-switching}.
One common approach is latent state machines. For example, HMM \citep{rabiner1986an}, RNN \citep{bengio1994learning} and their variants
\citep{hochreiter1997long, chung2015gated} use series of latent representations to encode temporal data.
However, such black-box encoding does not directly capture representative patterns (or referred to as ``states") that carry physical meanings in practice, such as walk or run in the observations from fitness-tracking devices.
While these methods sometimes can obtain strong results, they are still sensitive to noises \citep{senin2013sax-vsm}, provide poor interpretability, and are hard to debug when things go wrong.
For this reason, many recent studies focus on discretizing time-series and finding the underlying states, with methods such as sequence clustering \citep{hallac2017toeplitz, Yang2014HMM}, dictionaries (e.g. SAX \citep{senin2013sax-vsm,  lin2007experiencing}, BoP \citep{Lin2012Rotation}) and shapelets \citep{rakthanmanon2013fast, lines2012shapelet}. 
While effectively handling noises and providing better interpretability, they only recognize the states but \textit{ignore the potential effects of relations among them}.

\begin{figure}[t]
	\begin{minipage}{0.48\textwidth}
		\centering
		\includegraphics[width=0.9\textwidth]{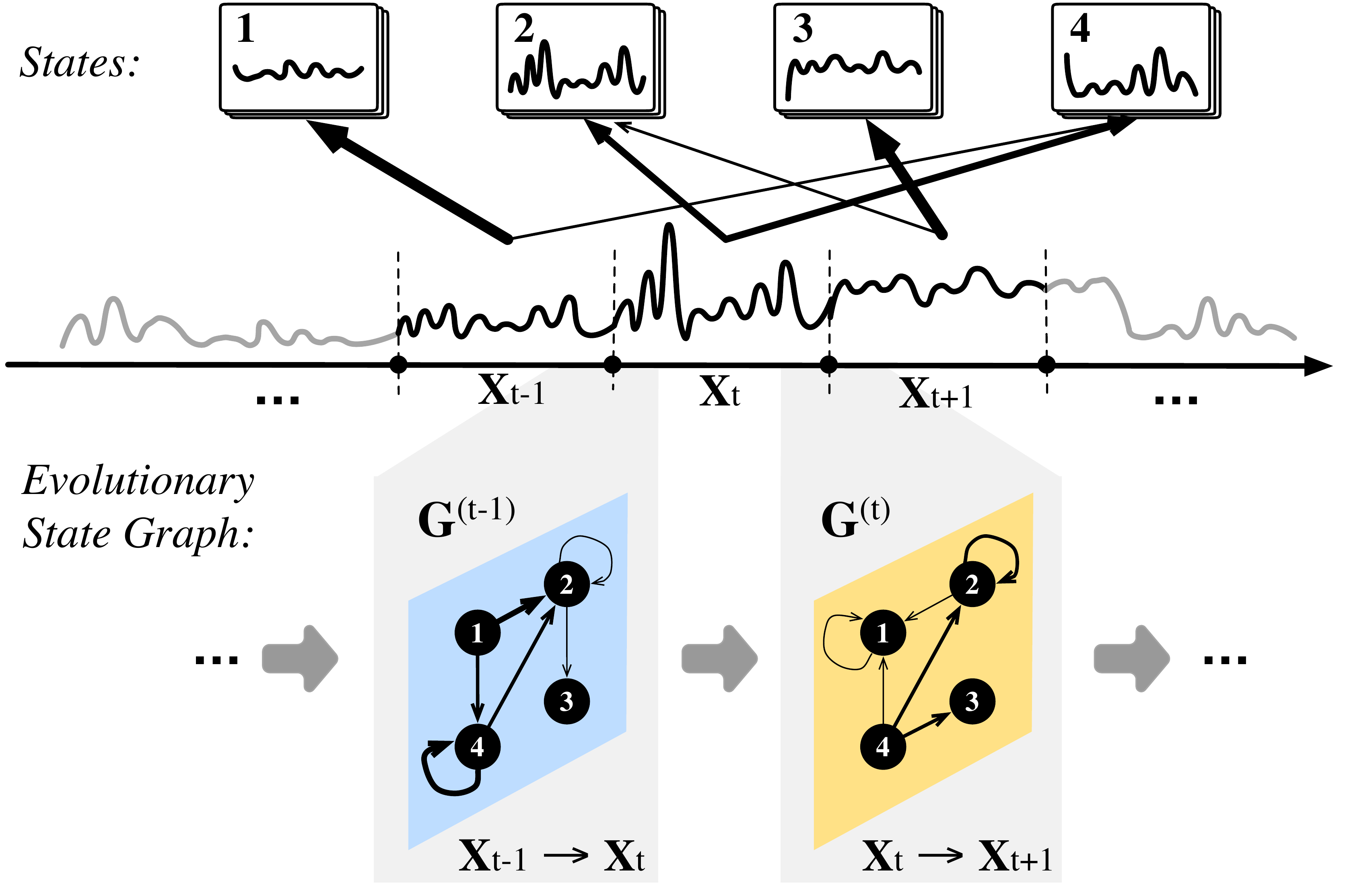}
		\caption{\footnotesize time-series can be segmented and recognized with several states (e.g., 1-4). Based on this, we construct the evolutionary state graph, where each node indicates a state and the edges represent their transitional relations across adjacent segments. Upon this, we develop \methodshort to further capture significant modes for effective time-series event prediction. \normalsize}
		\label{fig:intro}
	\end{minipage}
\end{figure} 

To jointly model the states and their relations, recent studies have started to explore the usage of \textit{graph structures}, such as GCN-LSTM \citep{liu2019characterizing} and Time2Graph \citep{cheng2019time2graph}.
However, GCN-LSTM requires an explicit graph as input (e.g., in-app action graph), which is difficult to directly get from general time-series data. 
Time2Graph uses shapelets to discover states and relations, but it only computes a single static graph over the whole timeline, despite the fact that the state relations might change over time (e.g., node-level dynamics and graph-level migration, cf.~\secref{sec:setup:observe} for details).
To the best of our knowledge, no existing studies have successfully captured and modeled the time-varying relations among the time-series states.

In this work, we observe that time-series are often affected by the joint influence of different states, and in particular, the \textit{change of relations among states}. 
For example, in the sequential observations from fitness-tracking devices, \textit{stopping exercise} from an \textit{intense run} may cause the \textit{fainting} event, while the monitoring data will look normal if one \textit{stops exercise} from \textit{jogging}; from online shopping records, a sudden interest change from \textit{electronics} to \textit{cosmetics} might be more suspicious than a smooth one from \textit{cosmetics} to \textit{fashion}.

Motivated by such observations, we propose a novel framework for time-series event prediction, by constructing and modeling a dynamic graph structure as shown in Figure \ref{fig:intro}.
Following existing studies \citep{Lin2012Rotation, senin2013sax-vsm, lines2012shapelet, cheng2019time2graph}, 
we model time-series based on the underlying \textit{states}. 
However, to preserve more information from the original time-series data, we model each \textit{time-series segment} as belonging to multiple states with different \textit{recognition weights}, and leverage a directed graph to model the \textit{transitional relations} among states between adjacent segments.
Since the graph evolves along the time-series, we refer to it as an \textit{\graphname}. 
Our empirical observations find that: 1) time-series evolution can be translated into different levels of graph dynamics; 2) when an event occurs, the time-series fluctuation can be expressed as the migration of graph structure, in particular, the dynamics of some edges connecting certain states (\secref{sec:setup:graph}).

Despite the insights provided by our empirical observations, there still remains the challenge of how to quantitatively leverage the \graphname to improve the performance of time-series event prediction. 
Existing GNN models only consider a static graph or node-level dynamics \citep{pareja2019evolvegcn, cheng2019time2graph, liu2019characterizing, li2016gated}, which cannot be directly used for learning with our \graphname.
In light of this, we propose a novel GNN model, \textit{\methodname} (\methodshort), to further model the graph-level propagation and node-graph interactions with a temporal attention mechanism.
The learned representations are then fed into an end-to-end model for time-series event prediction (\secref{sec:model:net}).

To validate the effectiveness of \methodshort, we conduct experiments on five real-world datasets. 
Our experimental results demonstrate the superiority of \methodshort over 11 state-of-the-art baselines on time-series event prediction (\secref{sec:exp:clf}). 
We further conduct comprehensive ablation and hyper-parameter studies to validate the effectiveness of our proposed method (\secref{sec:exp:pa}). 
Finally, we demonstrate the insights towards prediction explanation by visualizing \methodshort and its \graphname (\secref{sec:exp:case}).

The main contributions of this work are summarized as follows:
\begin{itemize} [leftmargin=*]
	\item Through real-world data analysis, we find the time-varying relations among states important for time-series event prediction.
	\item We propose the \graphname to capture the dynamic relations among states, and develop \methodshort to improve the performance of event prediction based on such graphs.
	\item We conduct extensive experiments on five datasets to demonstrate that our method can both make more accurate predictions, and provide more insight towards explaining them.
\end{itemize}

%% file: setup.tex

\section{Background and Problem}
\label{sec:setup}

\vpara{\textit{Time-series event prediction}.}
We consider the task of predicting future events in a given time-series sequence, following similar definition in previous work \citep{du2016recurrent, ning2016modeling, liu2019characterizing, ailliot2012markov-switching}.
Each time-series sequence with $T$ chronologically paired segments can be represented as
$$
\setlength\abovedisplayskip{1pt}
\setlength\belowdisplayskip{1pt}
\langle \mathbf{X}_{1:T}, \mathbf{Y}_{1:T} \rangle = \left\{\left(\mathbf{X}_1, \mathbf{Y}_1\right), \left(\mathbf{X}_2, \mathbf{Y}_2\right),...,\left(\mathbf{X}_T, \mathbf{Y}_T\right)\right\},
$$
where $\mathbf{X}_t \in \mathbb{R}^{\tau \times d}$ and $\mathbf{Y}_t \in \mathbb{Z}$ denote a \textit{time-series segment} \citep{bagnall2017the} and the \textit{observed event} in the corresponding time (e.g., anomalies), respectively.
Each segment $\mathbf{X}_t$ is a contiguous subsequence, i.e., $\mathbf{X}_t = \left\{\mathbf{x}_1,...,\mathbf{x}_{\tau}\right\}$, where $\mathbf{x}_i \in \mathbb{R}^d$ is a $d$-dimensional observation at the $i$-th time unit;  segment length $\tau$ is a hyper-parameter which indicates certain physical meanings (e.g. 24 hours).
If a time-series sequence can be divided by $T$ segments of equal length $\tau$, we then have 
$\langle \mathbf{X}_{1:T}, \mathbf{Y}_{1:T} \rangle= \left\{\left(\{\mathbf{x}_{\tau\times t+1},...,\mathbf{x}_{\tau\times t+\tau}\}, \mathbf{Y}_t\right)_{0 \leq t < T} \right\}$.
In this work, we aim to predict the future event $\mathbf{Y}_{T+1}$ via discovering time-series states behind $\langle \mathbf{X}_{1:T},  \mathbf{Y}_{1:T} \rangle$ and modeling their dynamic relations. 

\vpara{\textit{State}. }
\label{sec:setup:state}
A \textit{state} $v$ is a segment that indicates a representative pattern in the time-series sequence, denoted as $\bm{\Theta}_v\! \in\! \mathbb{R}^{\tau \times d}$. 
In our study, we adopt existing methods (e.g., Symbolic Aggregate Approximation \citep{senin2013sax-vsm,  lin2007experiencing}, Bag of Patterns \citep{Lin2012Rotation}, Shapelets \citep{rakthanmanon2013fast, lines2012shapelet}, sequence clustering \citep{hallac2017toeplitz, Yang2014HMM}) for recognizing interpretable states from time-series data (e.g., symbolic values, shapes or clusters), which are shown to be effective in handling noises and providing good interpretability.
As a minor but necessary contribution, we present different implementations of state recognition in the appendix (\secref{sec:appendix:state}), which act as interchangeable data pre-processors in our framework, and we conduct experiments in \secref{sec:exp:pa} to compare them.

\vpara{\textit{Segment-to-state representation}. }
Once the states have been recognized, one can then models each time-series segment $\mathbf{X}_t$ as a composition of states--i.e., quantify the \textit{recognition weight} of each state for a segment to characterize the segment-state associations.
Formally, given a segment $\mathbf{X}_t$ and a state $\bm{\Theta}_v$, the recognition weight $\mathbf{P}(\bm{\Theta}_v | \mathbf{X}_t)$ is a measurement of similarity, defined as follows.
\small
\begin{equation}
\setlength\abovedisplayskip{1pt}
\setlength\belowdisplayskip{1pt}
\mathbf{P}(\bm{\Theta}_v | \mathbf{X}_t) =  \frac{\max([\mathcal{D}(\mathbf{X}_t, \bm{\Theta}_v)]_{v \in \mathcal{V}}) - \mathcal{D}(\mathbf{X}_t, \bm{\Theta}_v)} {\max([\mathcal{D}(\mathbf{X}_t, \bm{\Theta}_v)]_{v \in \mathcal{V}}) - \min([\mathcal{D}(\mathbf{X}_t, \bm{\Theta}_v)]_{v \in \mathcal{V}})},
\label{eq:state:define}
\end{equation}
\normalsize
where $\mathcal{D}(\mathbf{X}_t, \bm{\Theta}_v)$ can be formalized as the Euclidean Distance or other distances based on different time-series representation and state recognition methods. (cf. \secref{sec:appendix:state} for details in the appendix).
The smaller this distance, the higher the weight $\mathbf{P}(\bm{\Theta}_v | \mathbf{X}_t)$. 

%% file: model.tex

\section{\methodshort Framework}
\label{sec:model}
In this section, we present a novel framework for time-series event prediction. We name the proposed framework \textit{\methodname} (\textit{\methodshort}), as it transforms the time-series into a dynamic graph based on the states and recognition weights, and constructs a GNN-based neural network to capture significant correlations and improve the ability of event prediction. 

\begin{figure*}[t]
	\begin{minipage}{1.\textwidth}
		\centering
		\includegraphics[width=0.9\textwidth]{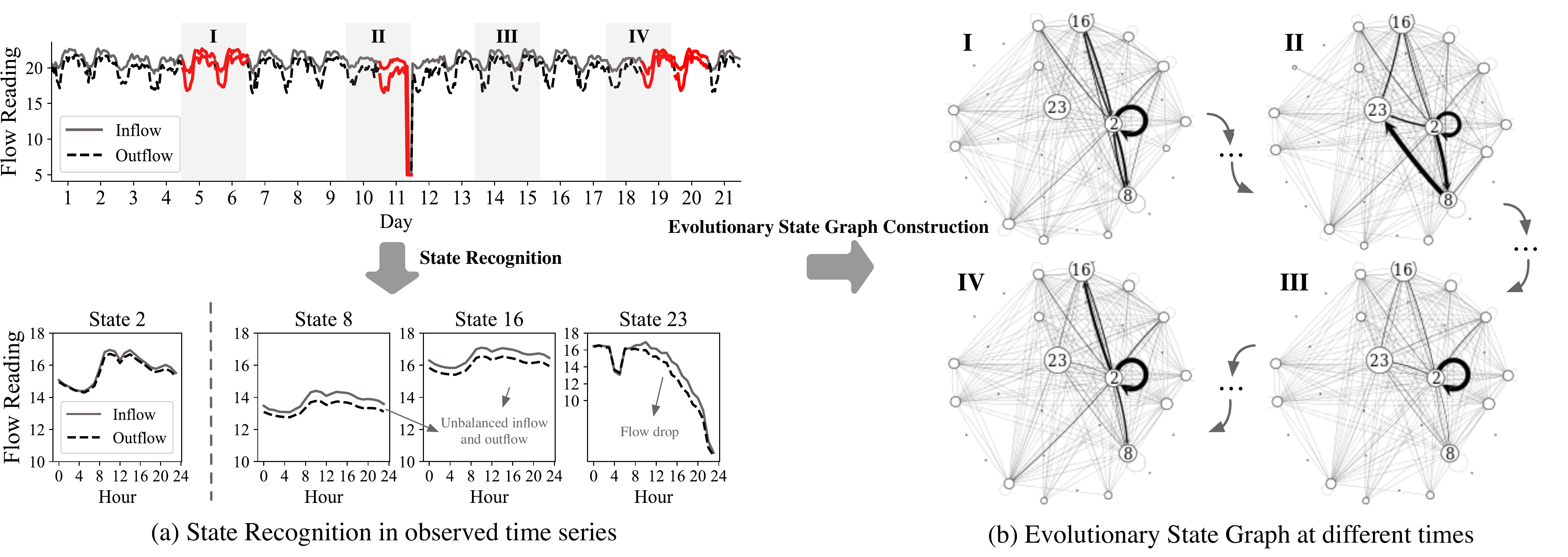}
		\caption{\small Example of an \textit{\graphname} constructed to predict network anomalies. \footnotesize (a)-(b) present a case in the \textit{NetFlow} dataset, such that the hourly inflow and outflow are recorded by the network monitor, while the red line indicates an anomaly occurred in this day. 
			(a) visualizes four states recognized by \methodshort, while (b) presents the \graphname in four different intervals (I, II, III, IV marked in (a)).
			Each node in the graph indicates a state, while edge's thickness indicates the weights of relations. 
			The thicker the edge, the greater the weight.
			We can see that the \graphname can help to derive insights to analyze time-series, 
			such as the fact that an unbalanced inflow and outflow or flow drop will lead to anomalies (e.g., state transitions: \#2$\rightarrow$\#16, \#8$\rightarrow$\#23). 
			\normalsize}
		\label{fig:observe:netflow:case}
	\end{minipage}
\end{figure*} 

\begin{figure}[t]
	\begin{minipage}{0.48\textwidth}
		\centering
		\includegraphics[width=1.0\textwidth]{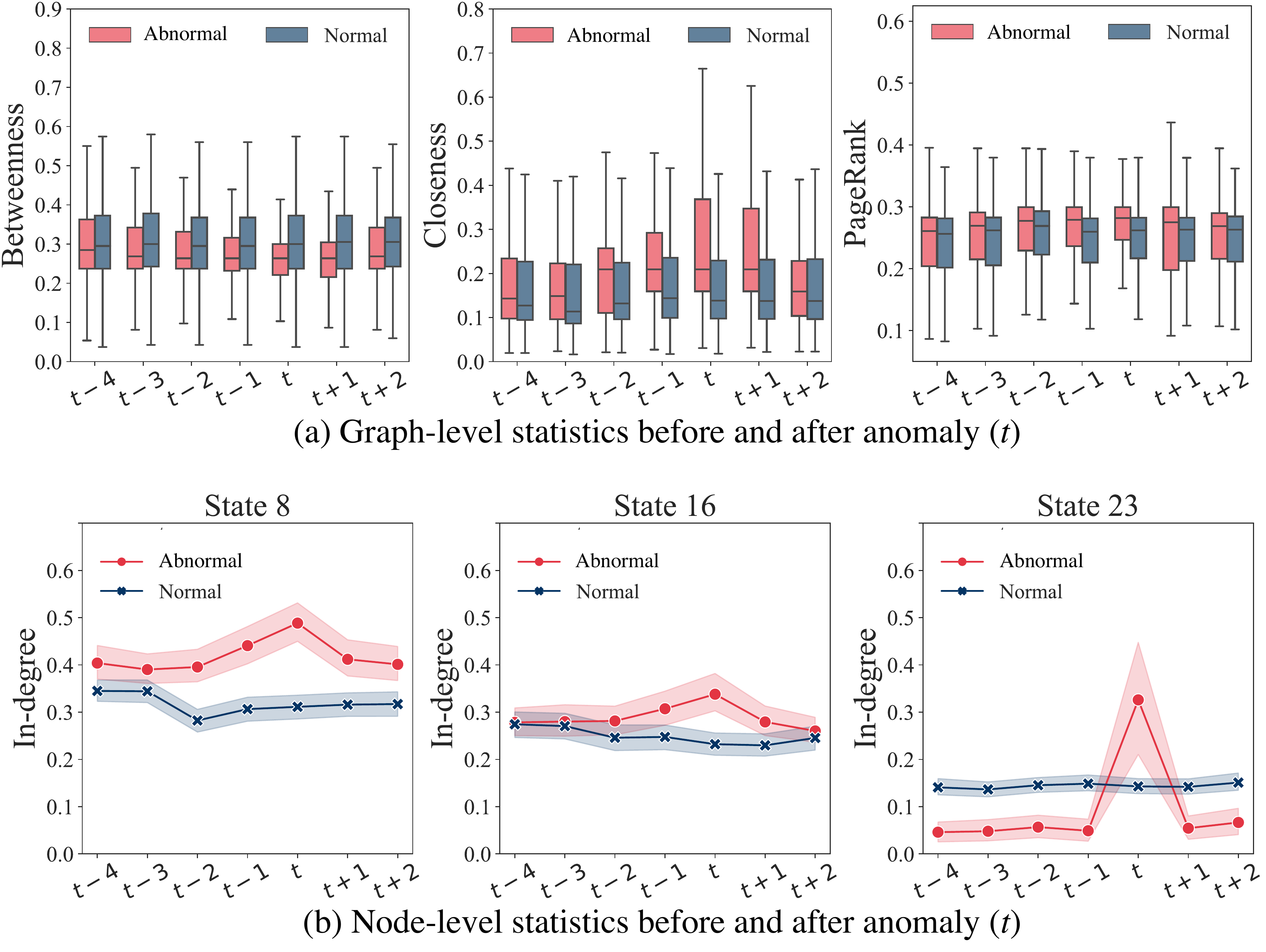}
		\caption{\small The statistics of the constructed \graphname  in \figref{fig:observe:netflow:case}. 
			\footnotesize (a)-(b) present some statistics based on the graph-level measurements (e.g., betweenness \citep{brandes2001a}, closeness \citep{opsahl2010node}, pagerank \citep{page1999the}) and node-level ones (in-degree) before and after anomaly $t$.
			\normalsize}
		\label{fig:observe:netflow:statistics}
	\end{minipage}
\end{figure} 

\vspace{-0.2cm}
\subsection{Evolutionary State Graph}
\label{sec:setup:graph}
Inspired by existing models introduced in \secref{sec:setup}, we aim to leverage the underlying states for effective and interpretable modeling of time-series.
A straightforward approach is to regard a time series as a sequence of the \textit{most likely states} (for each segment in the sequence), and then model their sequential dependencies \citep{hu2019capturing, hallac2017toeplitz, ailliot2012markov-switching}.
However, one segment may not belong only to a single state; rather it should be recognized as multiple states with different weights.
To this end, one can adopt a multiscale recurrent network (\textit{MRNN}) \citep{pascanu2013on} to model a multidimensional sequence of state weights, but this method does not highlight the \textit{transitions among the states}, which may essentially determine whether an event occurs.
Therefore, in this work we propose a novel dynamic graph structure to describe the relations among the states and explore how the dynamic shifts of states can reveal time-series evolution.

\vpara{\textit{Evolutionary state graph}.} 
We define the \textit{\graphname} as a sequence of weighted-directed graphs $\langle\mathbf{G}^{(1:T)}\rangle$. 
Specifically, each graph is formulated as $\mathbf{G}^{(t)}\! = \!\{\mathcal{V}, \mathcal{E}^{(t)}, \mathcal{M}^{(t)}\}$ to represent the transitions from the states of segments $\mathbf{X}_{t-1}$ to those of $\mathbf{X}_{t}$.  
Each node in the graph indicates a state $v$; each edge $e_{(v, v')}^{(t)} \in \mathcal{E}^{(t)}$ represents the \textit{transitional relation} (or \textit{relation} in short) from $v$ to $v'$, along with the \textit{transition weight} $m_{(v, v')}^{(t)} \in \mathcal{M}^{(t)}$.
Assuming the state weights observed for each segment to be independent, the transition weights are computed by
\begin{equation}
\setlength\abovedisplayskip{1pt}
\setlength\belowdisplayskip{1pt}
m_{(v, v')}^{(t)} = \mathbf{P}\left(\bm{\Theta}_{(v, ~v')}| \mathbf{X}_{(t-1, ~t)}  \right) = \mathbf{P}\left(\bm{\Theta}_{v} | \mathbf{X}_{t-1} \right) \times \mathbf{P}\left(\bm{\Theta}_{v'} | \mathbf{X}_{t} \right),
\label{eq:graph:build}
\end{equation}
which is the joint weight that $\mathbf{X}_{t-1}$ is recognized to the state $v$, while $\mathbf{X}_{t}$ is recognized to the state $v'$.

Compared with existing time-series representations based on states, our \graphname preserves more information from the original data along the timeline through the modeling of multiple states in each segment and their changing transitional relations. It allows the subsequent model to be more powerful and provide richer interpretations in its predictions, while inheriting from state-based representations the robustness towards noises.

\begin{figure*}[t]
	\begin{minipage}{1\textwidth}
		\centering
		\includegraphics[width=0.9\textwidth]{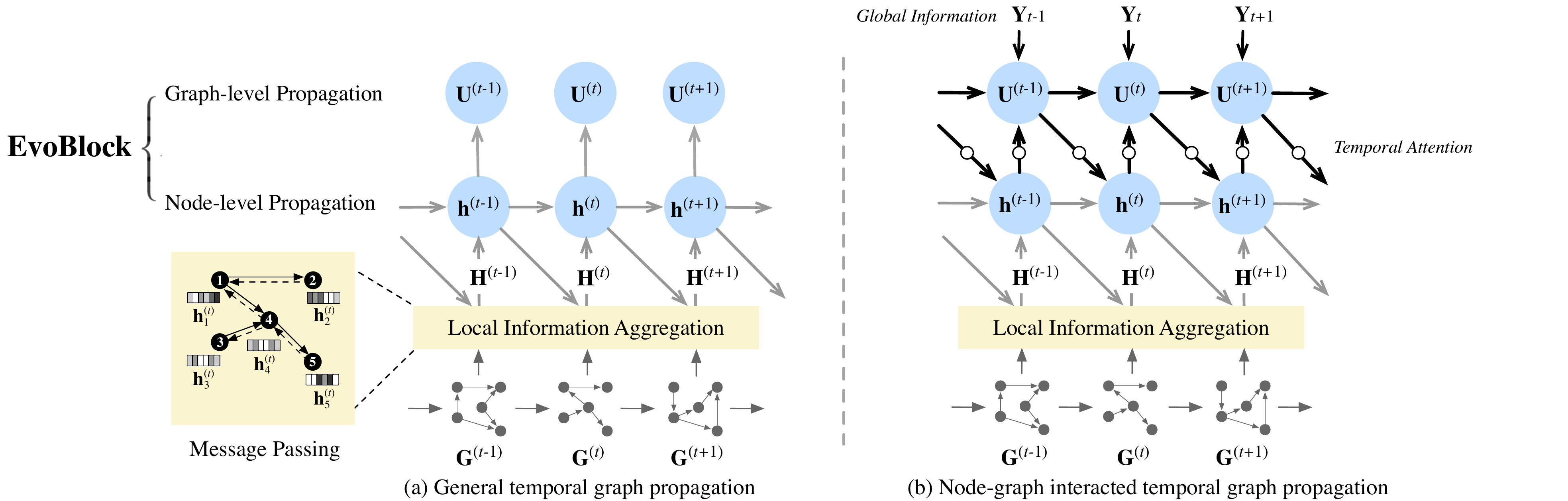}
		\caption{\small Overview of \methodshort.
			\footnotesize  After building the \graphname on the raw time series, 
			\methodshort conducts \textit{local information aggregation} and \textit{temporal graph propagation}. 
			For local information aggregation, each node in graph $\mathbf{G}^{(t)}$ has a feature vector $\mathbf{h}_v^{(t)}$. The solid edges indicate the passing messages, while the dashed edges indicate the feedback ones. 
			The graph-level patterns $\mathbf{U}^{(t)}$ and node-level features $\mathbf{h}^{(t)}$ are then propagated by the recurrent \textit{EvoBlock}, based on the aggregated intermediate representation $\mathbf{H}^{(t)}$:
			a) The general architecture for \textit{EvoBlock} on the \graphname, where $\mathbf{U}^{(t)}$ is pooled from each $\mathbf{h}^{(t)}_v, v\in\mathcal{V}$; b) The architecture of \methodshort, where graph-level and node-level propagation influence each other, based on the temporal attention mechanism.
			\normalsize}
		\label{fig:model}
	\end{minipage}
\end{figure*}

\vpara{Real-world example and analysis of \graphname.}
\label{sec:setup:observe}
To demonstrate how the \graphname reveals the evolution of time-series and helps the prediction of events, we conduct an observational study on the \textit{Netflow} dataset (cf. \secref{sec:exp:data} for details of the dataset).
As we can see from the case shown in \figref{fig:observe:netflow:case}, when an anomaly event occurs, the state transitions (\#2$\rightarrow$\#16) and (\#2$\rightarrow$\#8) are more frequent at time \uppercase\expandafter{\romannumeral1}; similarly, the state transitions (\#2$\rightarrow$\#8) and (\#8$\rightarrow$\#23) are obvious at time \uppercase\expandafter{\romannumeral2}. 
These transitions reveal that the unbalanced inflow and outflow (state 8 and state 16),  or flow drop (state 23), will cause anomalies of network devices. 
At time \uppercase\expandafter{\romannumeral3}, no anomaly occurs during this period. We can see that states primarily stay in \#2. 
There is then an anomaly in the next immediate moment at time \uppercase\expandafter{\romannumeral4}.  
Accordingly, we can see a clear increase of the state transition \#2$\rightarrow$\#16.

In light of the observations, we conduct statistic analysis related to the constructed \graphname based on the abnormal samples (an anomaly occurs at time $t$) and normal samples (no anomaly occurs). 
The distributions of the different graph-level and node-level measurements at different times (before and after anomaly $t$) are visualized in \figref{fig:observe:netflow:statistics}. 
From the figure, we can clearly see that when an anomaly occurs, the abnormal graph (red bar) tends to be denser; i.e., the betweenness scores gets lower, while the closeness scores gets higher. 
\figref{fig:observe:netflow:statistics}b presents three typical states and compares their in-degree before and after anomaly $t$. We can see that the in-degrees of state 8 and 16, indicating the unbalanced inflow and outflow, gradually increase before $t$; this illustrates that the network gradually becomes abnormal. The in-degree of state 23 suddenly increases, indicating that the flow drop is an unexpected event.
When no anomaly occurs, we can see that the normal \graphname (blue lines) generally remains unchanged.

Through the example, we show how the transformation of time-series into evolutionary state graphs allows us to capture the relations between states and their evolution. Meanwhile, we also learn that the graph-level and node-level evolutions can reveal different contextual information related to the time-series events: the node-level evolution reveals the states' skips when events occur, while the graph-level evolution presents the time-series migration.
Intuitively, we shall capture these two levels of information simultaneously when learning with evolutionary state graphs.

\subsection{\methodname}
\label{sec:model:net}
\vpara{Overview. }
Motivated by \secref{sec:setup:graph}, unlike most existing works \citep{senin2013sax-vsm, lines2012shapelet, ailliot2012markov-switching} 
which model the independent effects of each state, we develop \methodshort to capture the following two types of information through the leverage of our \graphname:  

\begin{itemize} [leftmargin=*]
	\item \textit{Local structural influence}: the same state $v$ will cause different observations when $v$ is transmitted from different states. In other words, the relations among states matter. For example, \textit{stopping exercise} from an \textit{intense run} may cause fainting, while the monitoring data will look healthier if one \textit{stops exercise} from \textit{jogging}. 
	\item \textit{Temporal influence}: previous transitions of states will influence the current observed data. 
	For example, $($\textit{intense run} $\rightarrow$ \textit{jogging} $\rightarrow$ $\cdots$ $\rightarrow$ \textit{stopping exercise}$)$ and $($\textit{jogging} $\rightarrow$ \textit{jogging} $\rightarrow$ $\cdots$ $\rightarrow$ \textit{stopping exercise}$)$ lead to different fitness effects. 
\end{itemize}
The above two types of influence can be naturally represented by the \graphname: 
the local structural influence is primarily determined by local-pairwise relations among nodes in each graph, while the temporal influence is determined by how relations evolve over different graphs. Inspired by Graph Neural Networks (GNN) \citep{battaglia2018relational}, we model both the structural and temporal influences of \graphname by designing two mechanisms: \textit{local information aggregation} and \textit{temporal graph propagation}. 

\figref{fig:model} illustrates the overall structure of \methodshort. Given the observations $\langle\mathbf{X}_{1:T}, \mathbf{Y}_{1:T}\rangle$, we first recognize states for each segment $\mathbf{X}_t$ and construct the \graphname $\langle\mathbf{G}^{(1:T)}\rangle$. 
Next, we define a representation vector $\mathbf{h}_v^{(t)} \in \mathbb{R}^{|\mathbf{h}|}$ for each node $v$ in graph $\mathbf{G}^{(t)}$ to encode $v$'s node-level patterns, and define a representation vector $\mathbf{U}^{(t)}\! \in\! \mathbb{R}^{|\mathbf{U}|}$ 
for $\mathbf{G}^{(t)}$ 
to encode 
the graph-level information.
Based on this, \methodshort aggregates local structural information by means of message passing, and further incorporates temporal information using the recurrent \textit{EvoBlock}. \methodshort then applies the learned representations $(\mathbf{h}, \mathbf{U})$ towards the prediction task. 

\vpara{Local information aggregation. }
\label{sec:model:local}
In order to aggregate the local structural information in each $\mathbf{G}^{(t)}$, \methodshort aims to make two linked nodes share similar representations.
To achieve this,  we let each node representation $\mathbf{h}_v^{(t)}$  in $\mathbf{G}^{(t)}$ aggregate the \textit{messages} of its neighbors, and thus compute its new representation vector.
Initially, we let $\mathbf{h}_v^{\!(0)\!}\!=\!\bm{\Theta}_v$. 
Recall that $\bm{\Theta}_v$ is obtained from the state recognition on all segments, which records the time-series information of state $v$. 
Then, following the message-passing neural network (MPNN)~\citep{gilmer2017neural} directly, we have the following aggregation scheme: 
\begin{equation}
\setlength\abovedisplayskip{3pt}
\setlength\belowdisplayskip{1pt}
\mathbf{H}_v^{(t)} = \sum_{v' \in N(v)} \mathcal{F}_{\rm{MP}}\left(\mathbf{h}_{v'}^{(t-1)}, e_{(v, v')}^{(t)} \right)
\label{eq:aggregation}
\end{equation}
where $\mathbf{H}_v^{(t)}$ is the intermediate representation of node $v$ following aggregation, which combines the messages from all neighbors $N(v)$ in the graph $\mathbf{G}^{(t)}$. 
The message function $\mathcal{F}_{\rm{MP}}(\cdot, \cdot)$ can be implemented by many existing neural networks, such as GGNN\citep{li2016gated}:
\begin{equation}
\setlength\abovedisplayskip{1pt}
\setlength\belowdisplayskip{1pt}
\mathcal{F}_{\rm{MP}}\left(\mathbf{h}_{v'}^{(t-1)}, e_{(v, v')}^{(t)} \right)= W_{\rm{MP}} \cdot\left[m_{(v, v')}^{(t)} \times\mathbf{h}_{v'}^{(t-1)} \right] + b_{\rm{MP}}
\label{eq:edgerelation}
\end{equation}
\noindent where $m_{(v, v')}^{(t)} \times \mathbf{h}_{v'}^{(t-1)}$ is the passing message, while $W_{\rm{MP}}$ and $b_{\rm{MP}}$ are the learnable parameters, indicating the passing weight and bias. 
We also have other implementations for $\mathcal{F}_{\rm{MP}}$, such as pooling, GCN \citep{duvenaud2015convolutional}, GraphSAGE \citep{hamilton2017inductive}, GAT \citep{gat2018petar}, etc. (cf. \secref{sec:appendix:gnn} for details in the appendix).
Herein, we serve $\mathcal{F}_{\rm{MP}}$ as interchangeable modules in \methodshort and conduct experiments in \secref{sec:exp:pa} to analyze the effectiveness of different implementations.

\vpara{Temporal graph propagation.}
\label{sec:model:temporal}
In addition to aggregating the local structural information, previous transitions also influence current representations. Moreover, when events occur, the modes of the graph-level and node-level evolution will change (\secref{sec:setup:observe}). Intuitively, we should capture these two kinds of temporal information simultaneously.
To achieve this, we design a recurrent block, named \textit{EvoBlock}, to capture the evolving information in the \graphname. EvoBlock combines the local aggregated representation $\mathbf{H}_v^{(t)}$ and the past representation $\left(\mathbf{h}_v^{(t-1)}, \mathbf{U}^{(t-1)}\right)$, 
formulated as
\begin{equation}
\setlength\abovedisplayskip{1pt}
\setlength\belowdisplayskip{0pt}
\mathbf{h}_v^{(t)}, \mathbf{U}^{(t)} := \mathcal{F}_{\rm{recur}} \left(\mathbf{H}_v^{(t)}, \mathbf{h}_v^{(t-1)}, \mathbf{U}^{(t-1)} \right) \quad {\rm{for}} ~ v \in \mathcal{V}
\end{equation}
where $\mathcal{F}_{\rm{recur}} $ indicates a recurrent function that allows us to incorporate information from the previous timestamp in order to update current representations.
When there are few messages from other nodes, i.e., $m_{(v', v)}^{(t)} \rightarrow 0$, $\left(\mathbf{h}_v^{(t)}, \mathbf{U}^{(t)}\right)$ will be more influenced by the previous $\left(\mathbf{h}_v^{(t-1)}, \mathbf{U}^{(t-1)}\right)$. Otherwise, the messages will influence current representations more. 

As shown in \figref{fig:model}a, most existing works implement $\mathcal{F}_{\rm{recur}} $ using simple recurrent neural networks on node-level propagation (e.g., GGSNN \citep{li2016gated} adopts GRU \citep{chung2015gated}, GCN-LSTM \citep{liu2019characterizing} adopts LSTM \citep{hochreiter1997long}, etc.). For the graph-level propagation $\mathbf{U}^{(t)}$, these methods simply pool the node-level representations, i.e., $\mathbf{U}^{(t)} = \sum_{v \in \mathcal{V}}\mathbf{h}_v^{(t)}$.
However, in our empirical observations (\secref{sec:setup:observe}), both the graph and nodes in the \graphname will present different temporal information when events occur. 
In order to improve the ability of event prediction, $\mathcal{F}_{\rm{recur}} $ shall consider the contextual information of previous events $\mathbf{Y}_{1:T}$ when modeling the graph-level propagation, and then influence the node-level representations via the node-graph interactions.
Accordingly, events are generally scattered in the timeline; thus, we propose a temporal attention mechanism for capturing significant temporal information in node-graph interactions. 
More specifically, as shown in \figref{fig:model}b, we have
\small
\begin{equation}
\setlength\abovedisplayskip{1pt}
\setlength\belowdisplayskip{1pt}
\mathcal{F}_{\rm{recur}}\! \left(\!\mathbf{H}_v^{(t)},\! \mathbf{h}_v^{(t-1)}, \!\mathbf{U}^{(t-1)} \!\right) \!= \!
\left\{
\begin{split}
&\mathbf{h}_v^{(t)} \!= \!\Phi_{\rm{h}} \left(\mathbf{h}_v^{(t-1)}, \!\mathbf{H}_v^{(t)}\! \oplus\! \alpha_t \!\mathbf{U}^{(t-1)} \!\right) \\
&\mathbf{U}^{(t)} \!= \!\Phi_{\rm{U}} \left(\mathbf{U}^{(t-1)}, \!\mathbf{Y}_t \!\oplus\! \alpha_t \sum_{v \in \mathcal{V}} \mathbf{h}_v^{(t)} \right)\\
&\alpha_t \!=\! {\rm{softmax}} \left( W_{\alpha} \!\left( \mathbf{U}^{(t-1)}\! \oplus\! \sum_{v\in \mathcal{V}}\mathbf{H}_{v}^{(t)}\! \right)\! \right)
\end{split}
\right.
\label{eq:hierarchical}
\end{equation}
\normalsize
where ``$\oplus$'' indicates the concatenation operator.
The current node-level representation is computed using the function $\Phi_{\rm{h}}(\cdot, \cdot)$, based on the past representations $\left(\mathbf{h}_v^{(t-1)}, \mathbf{U}^{(t-1)}\right)$ and current aggregations $\mathbf{H}_v^{(t)}$, while the current graph-level representation is computed by $\Phi_{\rm{U}}(\cdot, \cdot)$ based on the  past $\mathbf{U}^{(t-1)}$ and current event $\mathbf{Y}_t$, as well as all node representations $\mathbf{h}_v^{(t)}$.
The attention score $\alpha_t$ re-weights the node-graph interaction of the $t$-th temporal step, which is computed based on the concatenated patterns of $\mathbf{U}^{(t-1)}$ and all aggregations $\sum_{v\in \mathcal{V}}\mathbf{H}_{v}^{(t)} $, under the learnable weight $W_{\alpha}$. 
We use the softmax function to normalize $\alpha_t$ during different time steps.

Recurrent function $\Phi_{*}(\cdot, \cdot)$ smooths the two inputted vectors of each temporal step, and can be implemented using many existing approaches.
Herein, we provide an example of $\Phi_{\rm{h}}(\cdot, \cdot)$ implemented by LSTM. 
Formally, we have
\small
\begin{equation}
\setlength\abovedisplayskip{1pt}
\setlength\belowdisplayskip{1pt}
\begin{split}
&\Phi_{\rm{h}} \left(\mathbf{h}_v^{(t-1)}, \mathbf{H}_v^{(t)} \oplus \alpha_t \mathbf{U}^{(t-1)} \right) = \\ 
&\left\{
\begin{split}
\bm{\iota}^{(t)} \!&=\! \mathbf{H}_v^{(t)} \!\oplus\! \alpha_t \mathbf{U}^{(t-1)}  \\
\mathbf{F}^{(t)} \!&=\! \sigma(W_{\mathbf{F}} \!\cdot\! [\mathbf{h}_v^{(t-1)},\! \bm{\iota}^{(t)} \!] \!+\! b_{\mathbf{F}})  \quad
\mathbf{I}^{(t)} \!=\! \sigma(W_{\mathbf{I}} \! \cdot \!  [\mathbf{h}_v^{(t-1)}, \!\bm{\iota}^{(t)}\! ]\! +\! b_{\mathbf{I}}\!) \\
\mathbf{C}^{(t)}\! &=\! \mathbf{F}^{(t)} \!\circ\! \mathbf{C}^{(t-1)} \!+ \! \mathbf{I}^{(t)} \!\circ\! tanh(W_{\mathbf{C}} \!\cdot\! [\mathbf{h}_v^{(t-1)}, \!\bm{\iota}^{(t)} \!] \!+\! b_{\mathbf{C}}) \\
\mathbf{O}^{(t)} \!&=\! \sigma(W_{\mathbf{O}} \!\cdot\! [\mathbf{h}_v^{(t-1)}, \!\bm{\iota}^{(t)} \!] \!+\! b_{\mathbf{O}} \!)  \quad
\mathbf{h}_v^{(t)} \!=\! \mathbf{O}^{(t)}\! \circ\! tanh(\mathbf{C}^{(t)})
\end{split}
\right.
\end{split}
\label{eq:lstmlike}
\end{equation}
\normalsize
where $\mathbf{F}^{(t)}$, $\mathbf{I}^{(t)}$ and $\mathbf{O}^{(t)}$ are forget gate, input gate and output gate respectively, while $\sigma$ is a sigmoid activation function. 
The current node vectors are updated by receiving their own previous memory and current memory. 
In our experiments, we compare the performance of different methods for EvoBlock (\tableref{tb:exp:classresult}).

\vpara{End-to-End Model Learning.}
Thus, the representations $\mathbf{h}_v^{(t)}$ and $\mathbf{U}^{(t)}$ capture both the node-level and graph-level information respectively until the $t$-th temporal step, which can then be applied to predict the next event $\mathbf{Y}_{t+1}$.
More specifically, we encode the current evolutionary state graph $\mathbf{G}^{(t)}$ into representation $\mathbf{h}_{\mathbf{G}}^{(t)}$ based on the concatenated features of all $\mathbf{h}_{v}^{(t)}$ and $\mathbf{U}^{(t)}$, which can be formulated as 
\small
\begin{equation}
\setlength\abovedisplayskip{0pt}
\setlength\belowdisplayskip{0pt}
\mathbf{h}_{\mathbf{G}}^{(t)} = \mathcal{F}_{\rm{fc}} \left( \mathbf{U}^{(t)} \oplus \sum_{v\in \mathcal{V}}\mathbf{h}_{v}^{(t)}  \right)
\label{eq:output}
\end{equation}
\normalsize
where $\mathcal{F}_{\rm{fc}}$ acts as a fully connected layer. We then learn a classifier, such as a neural network or XGBoost \citep{chen2016xgboost}, which takes $\mathbf{h}^{(t)}_{\mathbf{G}}$ as input and estimates the probability of the next event, $\mathbf{P}\left(\mathbf{Y}_{t+1} |~ \mathbf{h}_{\mathbf{G}}^{(t)}\right)$. 
To learn the parameters $\theta$ of the proposed \methodshort and classifier, we employ an end-to-end framework, based on the Adam optimization algorithm \citep{kingma2015adam} to minimize the cross-entropy loss $\mathcal{L}$ as follows:
\small
\begin{equation}
\mathcal{L}= -\sum \hat{\mathbf{Y}}_{t+1}\log\mathbf{P}\left(\mathbf{Y}_{t+1} | \mathbf{h}_{\mathbf{G}}^{(t)}\right) \!
+\!(1 \!-\! \hat{\mathbf{Y}}_{t+1})\log \left(1 \!-\!\mathbf{P}\left(\mathbf{Y}_{t+1} | \mathbf{h}_{\mathbf{G}}^{(t)}\!\right) \!\right)
\label{eq:loss}
\end{equation}
\normalsize
where $\hat{\mathbf{Y}}_{t+1} \!\in\! \{0, 1\}$ is the ground truth that indicating whether a future event will occur.
The procedure of state recognition and graph propagation are carried out step by step: we first recognize the states and construct an evolutionary state graph, then conduct the \graphname propagation to model the time-series.
$|\mathcal{V}|$-node graphs are constructed in $T$ segments, such that the time complexity of each iteration is $O(T\times |\mathcal{V}|^2)$. 

%% file: exp.tex

\section{Experiments}
\label{sec:exp}
We apply our method to the prediction of upcoming events in time-series data, and aim to 
answer the following three questions:
\begin{itemize} [leftmargin=*]
	\item{\textbf{Q1:}} How does \methodshort perform on the time-series prediction task, compared with other baselines from the state-of-the-art? 
	\item{\textbf{Q2:}} How does the proposed EvoBlock effectively bridge the graph-level and node-level information over time? 
	\item{\textbf{Q3:}} How do different configurations, e.g., state number, segmentation length, implementation of state recognition and message passing, influence the performance?
\end{itemize}

\subsection{Datasets}
\label{sec:exp:data}
We employ five real-world datasets to conduct our experiments, 
including two public ones  (\textit{\kagglestocksshort} and \textit{\kaggletrafficshort}) from \textit{Kaggle}\footnote{An online community of data scientists and machine learners.}, and another three (\textit{\telecomshort}, \textit{\clockshort} and \textit{\aliyunshort}) provided by China Telecom\footnote{A major mobile service provider in China.}, State Grid\footnote{A major electric power company in China.} and 
Alibaba Cloud\footnote{The largest cloud service provider in Asia.}, 
respectively. \tableref{tb:exp:data} presents the overall dataset statistics.

\vpara{\kagglestocks (\kagglestocksshort).} This dataset comes from Kaggle.
It contains around 15K daily readings, each of which records four observations on a trading day:
three kinds of trade price and a trade number.
The task is to predict abnormal price volatility 
(variance greater than 1.0) 
in the next week (five trading days) based on the most recent records from the past year (50 weeks). In total, we identify around 12K normal cases and 3K abnormal ones. 

\vpara{\kaggletraffic (\kaggletrafficshort).} This dataset comes from Kaggle.
It contains around 3M daily readings, each of which records the number of views for a specific Wikipedia article. The task is to predict whether there will be a rapid growth 
(curve slope greater than 1.0) 
in the next month (30 days) based on the most recent records from the past 12 months. In total, we identify  around 900K  positive cases (rapid growth) and 2M negative ones.

\vpara{\telecomdataset (\telecomshort).} This dataset is provided by China Telecom. It consists around 238K hourly readings, each of which records the hourly in- and out-flow of network devices.  When an abnormal flow goes through the device ports, an alarm will be recorded. Our goal is to predict future anomalies (next day) based on records from the past 15 days. In total, we identify around 200K normal cases and 20K abnormal ones.

\vpara{\clockdataset (\clockshort).} This dataset is provided by the State Grid of China. It consists of around 6M weekly readings, each of which records the deviation time and delay of watt-hour meters. When the deviation time exceeds 120, the meter is marked as abnormal. Our goal is to predict anomalies in the next month based on records from the past 12 months. In total, we identify around 5M normal cases and 1M  abnormal ones.

\vpara{\aliyundataset (\aliyunshort).} This dataset is provided by 
Alibaba Cloud. 
It consists of around 12K server monitoring series, each of which records the minutely readings of different metrics (e.g., CPU, disk, memory, etc.). When a server fails to respond, 
the log will record the anomaly. Our goal is to predict anomalies in next 5 minutes based on records from the previous one hour. In total, we identify 11.8K normal cases and 0.2K abnormal ones.

\begin{table}[t]
	\centering
	\renewcommand\arraystretch{1.1}
	\addtolength{\tabcolsep}{-2pt}
	\caption{\small Dataset statistics}
	\label{tb:exp:data}
	\scalebox{0.85}{
		\begin{tabular}{l|ccccc}
			\hline
			\textbf{Dataset}  & \textbf{\kagglestocksshort} & \textbf{\kaggletrafficshort} &  \textbf{\telecomshort} &\textbf{\clockshort} & \textbf{\aliyunshort}\\ 
			\hline
			\#(samples) & 15,540 & 2,992,184 & 238,000 & 6,879,834 & 12,224 \\
			positive ratio(\%) &19.5 &28.2 & 8.6 & 14.9& 1.5 \\
			\hline 
		\end{tabular}
	}
	\normalsize
\end{table}

\begin{table*}[t]
	\centering
	\renewcommand\arraystretch{1.1}
	\addtolength{\tabcolsep}{-4pt}
	\caption{\small Comparison of prediction performance on five real-world datasets (\%). \footnotesize The \textbf{bold text} indicates the best performance among all methods, while the \underline{underline text} indicates the second-best performance.
		\normalsize}
	\label{tb:exp:classresult}
	\scalebox{0.88}{
		\begin{tabular}{m{1.7cm}|l|cc|cc|cc|cc|cc}
			\toprule
			\multicolumn{2}{c|}{\multirow{2}{*}{\textbf{\diagbox{Models}{Datasets}}}} & \multicolumn{2}{c|}{\textbf{\kagglestocksshort}} & \multicolumn{2}{c|}{\textbf{\kaggletrafficshort}} & \multicolumn{2}{c|}{\textbf{\telecomshort}} & \multicolumn{2}{c|}{\textbf{\clockshort}} & \multicolumn{2}{c}{\textbf{\aliyunshort}} \\
			
			\cline{3-12}
			\multicolumn{2}{m{2cm}|}{~} &F1-score & AUC &  F1-score& AUC & F1-score& AUC & F1-score& AUC&  F1-score& AUC \\
			\midrule
			
			\multirow{3}{1.7cm}{\textit{Feature-based models}} & BoP \citep{Lin2012Rotation}&24.92$\pm$0.40 &50.92$\pm$0.19&44.31$\pm$0.33 &66.87$\pm$0.09 & 54.01$\pm$0.89 &81.36$\pm$0.45& 60.01$\pm$0.49&85.20$\pm$0.38&42.59$\pm$0.60&70.22$\pm$0.37\\
			~&FS \citep{rakthanmanon2013fast}  &24.38$\pm$0.97 &50.55$\pm$0.42&43.89$\pm$0.76 &66.96$\pm$0.23& 52.84$\pm$1.63 &79.21$\pm$0.69& 58.34$\pm$0.83&84.32$\pm$0.71&46.95$\pm$0.91&72.04$\pm$0.56\\
			~&SAX-VSM \citep{senin2013sax-vsm}  &26.06$\pm$0.45 &51.42$\pm$0.20& 44.66$\pm$0.49 &67.63$\pm$0.15& 61.11$\pm$1.44&83.95$\pm$0.71& 62.44$\pm$0.65&85.97$\pm$0.64&47.98$\pm$0.75&73.88$\pm$0.49\\
			
			\midrule
			\multirow{3}{1.7cm}{\textit{Sequential models}}&S-HMM \citep{ailliot2012markov-switching}  &25.20 $\pm$0.48&51.14$\pm$0.20 &43.09$\pm$0.41&66.54$\pm$0.12& 58.05$\pm$0.87&81.89$\pm$0.49& 59.55$\pm$0.60 &84.99$\pm$0.61&48.71$\pm$0.60&73.65$\pm$0.38 \\
			~&MRNN \citep{pascanu2013on} &21.20$\pm$0.42&49.39$\pm$0.19&44.43$\pm$0.57&67.51$\pm$0.17&69.15$\pm$0.93&85.11$\pm$0.49&60.95$\pm$0.87&85.06$\pm$0.76&47.08$\pm$0.69&72.21$\pm$0.46\\
			~&HRNN \citep{chung2017hierarchical} &26.43$\pm$0.87&52.66$\pm$0.29&45.79$\pm$0.82&68.27$\pm$0.26&72.42$\pm$1.25&91.19$\pm$0.57&61.14$\pm$1.19&85.38$\pm$0.83&50.93$\pm$0.78&78.13$\pm$0.51\\
			
			\midrule
			\multirow{5}{1.7cm}{\textit{Graphical models}}&GGSNN \citep{li2016gated} &23.72$\pm$0.91&51.56$\pm$0.31&43.30$\pm$1.25&67.14$\pm$0.38&72.92$\pm$1.54&90.38$\pm$0.68&64.96$\pm$1.13&86.81$\pm$0.84&48.79$\pm$0.83&74.08$\pm$0.50\\
			~&GCN-LSTM \citep{liu2019characterizing} &25.76$\pm$0.85&52.66$\pm$0.30&45.67$\pm$0.90&68.15$\pm$0.29&75.05$\pm$1.38&91.43$\pm$0.60&65.65$\pm$1.04&87.03$\pm$0.78&50.95$\pm$0.80&78.14$\pm$0.50\\
			~&EvolveGCN \citep{pareja2019evolvegcn} &26.16$\pm$1.24&53.01$\pm$0.55&45.90$\pm$1.58&68.38$\pm$0.41&75.21$\pm$2.47&91.56$\pm$1.08&65.82$\pm$1.92&87.17$\pm$1.29&50.63$\pm$1.37&78.01$\pm$0.98\\
			~&ST-MGCN \citep{geng2019spatiotemporal} &26.93$\pm$0.97&53.39$\pm$0.39&45.96$\pm$0.91&68.74$\pm$0.27&77.79$\pm$1.40&91.95$\pm$0.64&66.61$\pm$1.11&87.78$\pm$0.83&\underline{51.21$\pm$0.85}&\underline{78.33$\pm$0.52}\\
			~&Time2Graph \citep{cheng2019time2graph} &26.50$\pm$0.91&53.28$\pm$0.39&\underline{46.03$\pm$1.12}&\underline{68.74$\pm$0.43}&76.94$\pm$1.83&91.61$\pm$0.64&67.01$\pm$1.46&88.00$\pm$1.23&50.50$\pm$1.02&77.87$\pm$0.98\\
			
			\midrule
			\multirow{3}{1.7cm}{\textit{Our models}} &\methodshort w/o G
			&25.81$\pm$0.80&52.67$\pm$0.33&45.66$\pm$0.85&68.45$\pm$0.38&74.92$\pm$1.42&91.40$\pm$0.63&65.71$\pm$0.99&87.10$\pm$0.80&50.89$\pm$0.80&78.10$\pm$0.50\\
			~&\methodshort w/o A &\underline{29.11$\pm$0.83}&\underline{54.47$\pm$0.37}&45.95$\pm$0.91&68.55$\pm$0.25&\underline{79.37$\pm$1.43}&\underline{92.45$\pm$0.66}&\textbf{69.21$\pm$1.17}&\textbf{89.92$\pm$0.80}&51.20$\pm$0.81&78.10$\pm$0.50\\
			~&\methodshort &\textbf{30.47$\pm$0.93}&\textbf{55.07$\pm$0.39}&\textbf{47.02$\pm$0.95}&\textbf{69.03$\pm$0.27}&\textbf{80.25$\pm$1.43}&\textbf{92.67$\pm$0.65}&\underline{68.62$\pm$1.21}&\underline{89.76$\pm$0.82}&\textbf{53.44$\pm$0.87}&\textbf{79.97$\pm$0.52}\\
			
			\bottomrule 
		\end{tabular}
	}
\end{table*}

\subsection{Baseline Methods}
\label{sec:exp:baseline}
We compare our proposed \methodshort with several groups of baselines: 

\vpara{Feature-based models.}
Several popular feature-based algorithms have been proposed for time-series analysis. In this paper, we choose some typical algorithms to compare with our model: Bag of Patterns (\textit{BoP}) \citep{Lin2012Rotation}, Vector Space Model using SAX (\textit{SAX-VSM}) \citep{senin2013sax-vsm} and Fast Shapelet (\textit{FS}) \citep{rakthanmanon2013fast}. These methods capture different state representations, which serve as features for event predictions.

\vpara{Sequential models.} 
Another typical group of algorithms interpret the time-series as a new sequence of states, and model their sequential dependencies. In this paper, we use several famous frameworks as baselines: switching-time-series model (\textit{S-HMM}) \citep{ailliot2012markov-switching} models the Markov dependencies of state sequences; multiscale recurrent neural network (\textit{MRNN}) \citep{pascanu2013on} takes the concatenated multi-source sequences (\small$\mathbf{X}_t \oplus \mathbf{Y}_t$\normalsize) as input and learns one latent representation for prediction; hierarchical recurrent neural network (\textit{HRNN}) \citep{chung2017hierarchical} captures more correlations between $\mathbf{X}_t$ and $\mathbf{Y}_t$, which conducts the same mechanism as Evoblock.

\vpara{Graph-based models.} Recently, many GNN-based works are proposed to model the (dynamic) graphs. In this paper, we choose several state-of-the-art algorithms as baselines to model the \graphname, and conduct the same approaches for event prediction as \methodshort: gated graph neural network (\textit{GGSNN})\citep{li2016gated} initializes the node vector $\mathbf{h}^{(0)}$ using a one-hot vector of the corresponding state; it conducts GGNN\citep{li2016gated} for local message passing and only adopts a GRU structure\citep{chung2015gated} for node-level propagation.
\textit{GCN-LSTM}\citep{liu2019characterizing}
uses states' patterns $\bm{\Theta}$ to initialize the node vector $\mathbf{h}^{(0)}$; it conducts GCN\citep{duvenaud2015convolutional} for local message passing and LSTM structure\citep{hochreiter1997long} for node-level propagation. 
\textit{EvolveGCN}\citep{pareja2019evolvegcn} is a dynamic graph neural network that builds a multi-layer framework to combine RNN and GCN; it also focuses on node-level propagation.
\textit{ST-MGCN}\citep{geng2019spatiotemporal} is a spatiotemporal multi-graph convolution network, in which $\mathbf{Y}_t$ serve as contextual information for propagation. It directly fuses the contextual information into node-level representations rather than learning graph-level representations and modeling the node-graph interactions.
\textit{Time2Graph}\citep{cheng2019time2graph} adopts shapelet to extract states; it aggregates the graphs at different times as a static graph and conduct DeepWalk\citep{perozzi2014deepwalk} to learn graph's representations, which then serve as features for event predictions.

\vpara{\methodshort variants.}
We also compare \methodshort with its derivatives by modifying some key components to see how they fare: 1) we sample the most possible state sequence (i.e., each segment is recognized with highest state weight) for each time-series, and directly use LSTM to model the new sequence without building and modeling the \graphname, denoted as \textit{\methodshort w/o G}; 2) we build \graphname for time-series but model it without conducting temporal attention mechanism, denoted as \textit{\methodshort w/o A}; 3) we conduct complete \methodshort for time-series modeling, denotes as \textit{\methodshort}.
Herein, \methodshort uses the state patterns $\bm{\Theta}$ to initialize node vector $\mathbf{h}^{(0)}$ and conducts graph-level and node-level propagation for $\langle\mathbf{G}^{(1:T)}\rangle$.
We implement state recognition and local message passing using Kmeans \citep{kanungo2002an}  and GGNN \citep{li2016gated}  respectively. We will study how different implementations influence the performance later in \secref{sec:exp:pa}.

\subsection{Implementation details}
\label{sec:sec:setup}
We conduct experiments on the five real-world datasets. 
We split the train/test set by 0.8 at the time line, such that preceding segments are used for training and the following ones are used for testing. We also split 10\% samples from train set as validation set in order to avoid overfitting.
We run all experiments on a single GPU with a batch size of 1000, and train our models for 100 iterations in total, starting with a learning rate of 0.001 and reducing it by a factor of 10 at every 20 iterations. 
Due to limit space, the hyperparameter settings of different methods are presented in the appendix (cf. \secref{sec:appendix:setup} for details in the appendix).

\subsection{Performance Comparison}
\label{sec:exp:clf}
We compare the performance of \methodshort and other baselines in order to answer \textbf{Q1}, and also conduct ablation studies to answer \textbf{Q2}. 
For the binary event prediction tasks, we use F1 score and AUC as our evaluation metrics, due to the unbalanced positive ratio. 
All reports are the average results of five times repeated experiments, along with their standard deviations (see details in \tableref{tb:exp:classresult}).

\vpara{1. Feature-based models vs. others.}
We observe that all feature-based methods perform poorly, because they only capture the states as features but ignore the influence of relations. \textit{FS} is unstable relatively and \textit{SAX-VSM} outperforms another two methods.
We note that other models capture the relations and outperform feature-based ones, demonstrating the significance of relation modeling.

\vpara{2. Sequential models vs. graphical models.}
We compare the sequential models with graphical models in order to present the effectiveness of different methodologies for relation modeling. Most graph networks outperform \textit{MRNN} and \text{S-HMM}, 
illustrating that modeling the dynamic relations of states is more significant compared to modeling their sequential dependencies. \textit{HRNN} effectively improves the performance and even beats some graph neural models on the \kagglestocksshort datasets, which suggest that we should try to capture the multi-level correlations in the temporal modeling. 

\vpara{3. Effectiveness of temporal modeling.}
For the temporal modeling on the \graphname, different graph neural models adopt different mechanisms. Due to the monotonous information expression of one-hot annotations, \textit{GGSNN} is not as good as the latter methods.
Accordingly, \textit{GCN-LSTM} utilizes more state information and performs better. 
\textit{EvolveGCN}, which builds multi-layer deep networks to combine RNN and GCN, is unstable. 
\textit{Time2Graph} models the aggregated static graph and ignores the temporal dependencies, which dose not outperform \textit{ST-MGCN} and \textit{our models}. 
As we expect, our proposed \methodshort model conducts the node-graph interaction during the temporal graph propagation, making it more suitable for the temporal modeling of the \graphname.

\vpara{4. Ablation study on propagation mechanism.}
As shown in \tableref{tb:exp:classresult} (\textit{Our models}), we attempt to validate the effectiveness of the proposed EvoBlock. We can see that, due to simple modeling on state sequence, \textit{\methodshort w/o G} performs poorly. 
When we build and model the \graphname for time-series (\textit{\methodshort w/o A}), the performances are improved with the information of node-graph interaction.
The temporal attention mechanism can capture the significant correlations during the temporal propagation, meaning that it outperforms other implementations as expected (\textit{\methodshort}). We present several cases in \secref{sec:exp:case} to support this conclusion.

\subsection{Parameter Analysis}
\label{sec:exp:pa}
\begin{figure}[t]
	\begin{minipage}{.48\textwidth}
		\centering
		\includegraphics[width=0.9\textwidth]{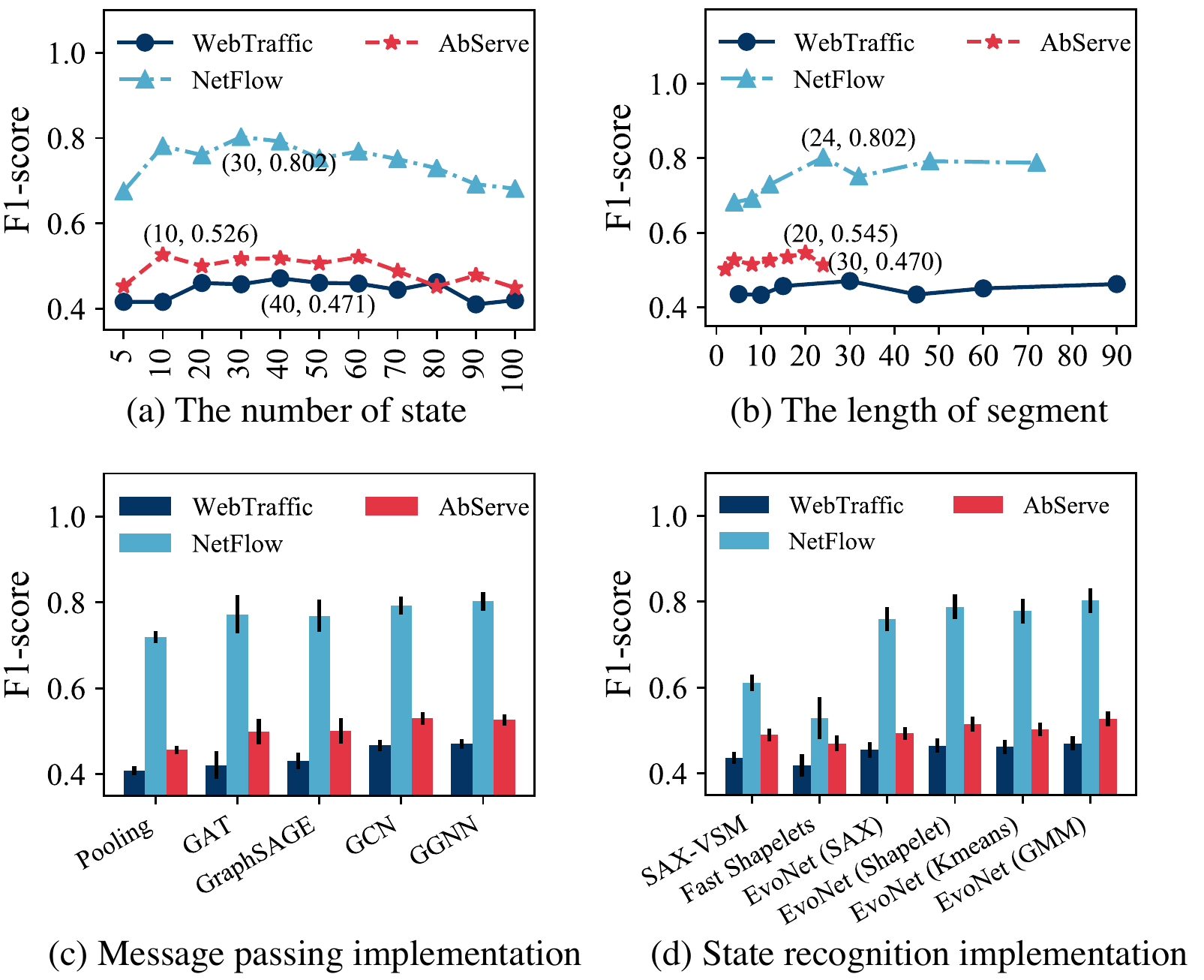}
		\caption{\small The impact of the different parameters. \footnotesize (a)-(d) present comparisons of state number $|\mathcal{V}|$, segment length $\tau$, implementation of message passing and state recognition, respectively, over three datasets (\kaggletrafficshort, \telecomshort and \aliyunshort) in \secref{sec:exp:data}. \normalsize}
		\label{fig:pa}
	\end{minipage}
\end{figure}

We examine the sensitivities of four important parameters to answer \textbf{Q3}: state number $|\mathcal{V}|$, segment length $\tau$, implementation of message passing and state recognition. Due to space limitations, we present the results based on only three datasets in \figref{fig:pa}. 
We test $|\mathcal{V}|$ with values from 5 to 100 with interval 10, and test $\tau$ with different lengths that are smaller or greater than the period length of the event.
We compare the pooling method, GAT \citep{gat2018petar}, GraphSAGE \citep{hamilton2017inductive}, GCN \citep{duvenaud2015convolutional} and GGNN \citep{li2016gated} for message passing, as well as SAX word \citep{senin2013sax-vsm}, Shapelets \citep{lines2012shapelet}, Kmeans \citep{kanungo2002an} and GMM \citep{bouttefroy2010on} for state recognition.
The F1-score is used as a metric to compare these parameters across the datasets.

\vpara{1. Sensitivities of state number $|\mathcal{V}|$.}
As shown in \figref{fig:pa}a, prediction performance curves differ depending on the dataset, illustrating that the state number $|\mathcal{V}|$ is sensitive to the data owns patterns. Moreover, the performance is not bound to improve as $|\mathcal{V}|$ increases, suggesting that $|\mathcal{V}|$ is an empirically determined parameter and is unsuitable for large values.

\vpara{2.  Sensitivities of segment length $\tau$.}
Another sensitive parameter is the segment length $\tau$, the variation of which may change the temporal scale of event $\mathbf{Y}_t$, and thus the positive ratio of ground truth.
We can see the performances in \figref{fig:pa}b do not vary significantly, meaning that it can be an empirical parameter that is generally determined by the realistic demand (e.g. an acceptable temporal scale of anomaly detection, etc.)

\vpara{3.  Implementation of message passing.}
\figref{fig:pa}c presents the comparisons for different implementations of message passing. We can observe that GAT and GraphSAGE perform poorly and are unstable due to their full attention or sampling operation, which is unsuitable for the small-scale graph. The performances of GGNN and GCN are similar, and both outperform the pooling method.

\vpara{4.  Implementation of state recognition.}
As shown in \figref{fig:pa}d, we test different implementations of state recognition, and further compare them with some feature-based baselines (i.e., SAX-VSM \citep{senin2013sax-vsm} and Fast Shapelets \citep{rakthanmanon2013fast}).
We can see that \methodshort can clearly improve the performance of SAX-VSM and Fast Shapelets when models the relations.
Moreover, the implementations of cluster methods and shapelet outperform the SAX word; this is because each SAX word is simply a symbolic value representing state, while other representations are a vector describing state patterns, which provide more information for modeling the \graphname.

\begin{figure}[t]
	\begin{minipage}{.48\textwidth}
		\centering
		\includegraphics[width=1.0\textwidth]{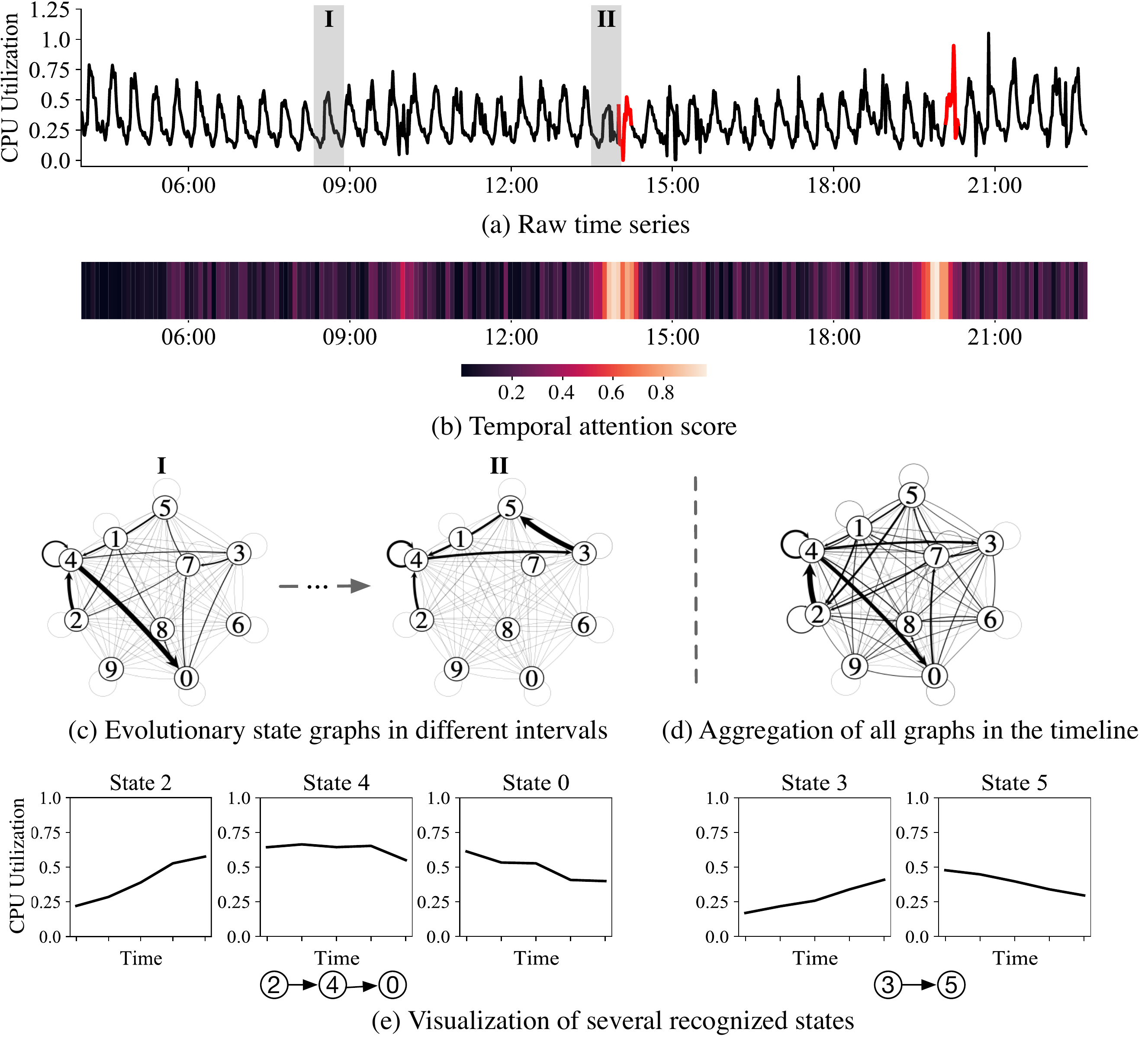}
		\caption{\small A case of \methodshort conducted for anomaly analysis of cloud service.
			\footnotesize (a) visualizes the raw time-series of CPU utilization. The red lines indicates anomaly events during different intervals ($\tau = 5$).
			We set 10 states for constructing \graphname and conducting \methodshort.
			Heat map in (b) presents the attention score $\alpha_t$ of \methodshort at different temporal steps. 
			(c)-(d) visualize the \graphname in two different intervals (I, II marked in (a)) and the aggregated graph in the whole timeline, respectively. 
			(e) presents the raw segments corresponding to the five different states.
			\normalsize}
		\label{fig:case:graph}
	\end{minipage}
\end{figure}

\subsection{Case Studies}
\label{sec:exp:case}
In this section, we apply our \methodshort method to a real-world anomaly prediction scenario in Alibaba Cloud\footnote{Our method has been deployed by \href{https://www.aliyun.com/product/sls}{SLS, Alibaba Cloud}, the largest log service provider in China, acting as a common function.},
enabling us to demonstrate how this method can be used to find meaningful relational clues to explain its results.
As described in \secref{sec:exp:data}, the minutely time-series of server monitor are segmented by the interval $\tau = 5$ (empirical length).
In order to present clearly, we cluster 10 states for constructing \graphname and conduct \methodshort for anomaly prediction. We visualize the results including several states and the \graphname at different times.
The temporal attention scores learned by \methodshort are also visualized to validate its effectiveness.
All results are presented in \figref{fig:case:graph}.

\vpara{1. Effectiveness of temporal attention mechanism.}
As shown in \figref{fig:case:graph}(a)-(b), we adopt heat map to visualize the attention scores $\alpha_t\!$ learned by \equationref{eq:hierarchical} at different times.
We can see that the attention scores successfully highlight the positions of anomalies in (a) (i.e., the positions near 13:00 and 20:00), which demonstrate that the temporal attention mechanism is useful for \methodshort to capture significant temporal information.

\vpara{2. Interpretability of \graphname.}
We then explore how an \graphname can be used to find meaningful insights that can explain anomaly event. 
As shown in \figref{fig:case:graph}(a), we mark two intervals, I and II, to visualize  \graphname and explore some meaningful insights. 
The results are shown in \figref{fig:case:graph}(c)-(d).
We can see that there is a major transition  \#2$\rightarrow$\#4$\rightarrow$\#0 (i.e., thick edges) in the graph of I, while \#3$\rightarrow$\#5 is a major transition in the graph of II.
Note that there is an anomaly occurring immediately after interval II.
When we aggregate all \graphnames in the timeline (\figref{fig:case:graph}(d)), we can find that the transition \#2$\rightarrow$\#4$\rightarrow$\#0 is the major path in the graph, while \#3$\rightarrow$\#5 is a rare path (i.e., thin edges). These observations indicate that the transition \#3$\rightarrow$\#5 occurred in interval II is abnormal, which is consistent with the anomaly of cloud service.
As shown in \figref{fig:case:graph}(e), we present the average curve of segments with different states.
We can see that the transition \#2$\rightarrow$\#4$\rightarrow$\#0 indicates a process of service, i.e., CPU utilization rises from 0.25 to 0.75 and drops after maintaining a period.
On the contrary, transition \#3$\rightarrow$\#5 indicates that CPU utilization rises to 0.5 and then drops immediately. 
These observations demonstrate that this anomaly may be caused by the CPU's fault.




%% file: related.tex

\section{Related Work}
\label{sec:related}
\vpara{time-series modeling.} time-series modeling aims to capture the representative patterns underpinning observed data. One important trend here is sequential modeling, such as HMM \citep{rabiner1986an}, RNN \citep{bengio1994learning} and their variants \citep{hochreiter1997long, chung2015gated, Yang2014HMM,hu2020theft} and fitting auto-regressive models \citep{Bagnall2014A}.
They define one latent representation to capture all the patterns by modeling the sequential dependencies, rather than distinguishing different states.
Another trend is mining discretized sequential patterns, such as switch time-series models \citep{ailliot2012markov-switching} and dictionaries \citep{senin2013sax-vsm, Lin2012Rotation, lin2007experiencing}. They model the time-series by capturing different states of segments independently, but ignore the influence from their relations.
Recently, some works use hierarchical or attention connections to incorporate the above two technique and get good performance \citep{wang2018multilevel, chung2017hierarchical}. However, most of them only capture the patterns of states, ignoring their relations.
Some works have applied \textit{graph structure} into the relation modeling of time-series states \citep{liu2019characterizing, cheng2019time2graph, hallac2017toeplitz}, which aims to represent different segments, rather than capturing the dynamics. To the best of our knowledge, no existing studies have successfully modeled the time-varying relations among states. 

\vpara{Graph neural networks.} Models in the graph family \citep{battaglia2018relational, gat2018petar, hamilton2017inductive, duvenaud2015convolutional, li2016gated, feng2018scalefree, zhou2018dynamic} have been applied to many real-world scenarios,
including learning the dynamics of physical systems \citep{battaglia2016interaction, sanchez2018graph}, predicting the chemical properties of molecules \citep{fout2017protein}, predicting traffic on roads \citep{geng2019spatiotemporal} and reasoning about knowledge graphs \citep{hamaguchi2017knowledge}, etc.. 
These studies present the effectiveness of GNNs for modeling structural information. 
Some works focus on summarizing models and refining formal expressions. The message-passing neural network (MPNN) unified various graph convolutional network and graph neural network approaches by analogy to message-passing in graphical models \citep{gilmer2017neural}. The non-local neural network (NLNN) has a similar vein, which unified various “self-attention”-style approaches by analogy to methods from graphical models and computer vision for capturing long range dependencies in signals \citep{wang2018non-local}.
Recently, some works have attempted to model dynamic graphs using GNNs \citep{pareja2019evolvegcn, liu2019characterizing, geng2019spatiotemporal}, although they focus primarily on the explicit graphical structure. 
To the best of our knowledge, no existing studies have successfully modeled dynamic relations in non-graphical data, such as time-series.

%% file: conclusion.tex

\section{Conclusions}
\label{sec:conclusion}
In this paper, we study the problem of how relations among states reflect the evolution of temporal data. 
We propose a novel representation, the \graphname, to present the time-varying relations among time-series states.
In order to capture these effective patterns for downstream tasks, we further propose a GNN-based model, \methodshort, to conduct dynamic graph modeling.
As for the validation of \methodshort's effectiveness, we conduct extensive experiments on five real-world datasets.
Experimental results demonstrate that our model clearly outperforms 11 state-of-the-art benchmark methods.
Based on this, we can find some meaningful relations among the states that allow us to understand temporal data.

%% file: appendix.tex

\section{Appendix}
\label{sec:appendix}

\subsection{Algorithm Details}
\label{sec:appendix:code}
In order to outline our proposed model in detail, we present the complete pseudo code of \methodshort to illustrate the learning procedure.
Given the observations $\langle \mathbf{X}_{1:T}, \mathbf{Y}_{1:T}\rangle$ and parameters $(|\mathcal{V}|, \tau, \mathcal{F}_{\rm{state}}, \\ \mathcal{F}_{\rm{MP}})$,  \methodshort first captures different states by means of the recognition function $\mathcal{F}_{\rm{state}}$. It then constructs the evolutionary state graph $\langle\mathbf{G}^{(1:T)}\rangle$ and conducts graph propagation by means of the message function $\mathcal{F}_{\rm{MP}}$ and EvoBlock. Finally, the learned representations are fed into an output model for prediction tasks; we use a back-propagation learning algorithm with cross-entropy loss to train the entire networks. More details can be found in \alref{alg:1}.

\begin{algorithm}[H]
	\renewcommand{\algorithmicrequire}{\textbf{Input:}}  
	\renewcommand{\algorithmicensure}{\textbf{Output:}}  
	\caption{The learning procedure of  \methodshort}
	\label{alg:1}
	\begin{algorithmic}[1]
		\Require Observations $\langle \mathbf{X}_{1:T}, \mathbf{Y}_{1:T}\rangle$, parameters $(|\mathcal{V}|, \tau, \mathcal{F}_{\rm{state}}, \mathcal{F}_{\rm{MP}})$
		\Ensure Model parameters $\theta$, event prediction $\mathbf{Y}'$ 
		
		\State $\mathcal{F}_{\rm{state}} \gets (\mathbf{X}_{1:T}, |\mathcal{V}|)$, train state recognition model
		\For{each segment $\mathbf{X}_t \in \mathbf{X}_{1:T}$}
		\State $\{\mathbf{P}(\bm{\Theta}_v | \mathbf{X}_t) \gets \mathcal{F}_{\rm{state}}(\mathbf{X}_{t}, \bm{\Theta}_{v})\}_{v \in \mathcal{V}}$, get the recognized weights of states as \equationref{eq:state:define}
		\State $\mathbf{G}^{(t)} \gets $ construct the \graphname as \equationref{eq:graph:build}
		\EndFor
		
		\\
		
		\State $\mathbf{U}^{(0)} \gets \bm{0}$ initialize the graph-level representation
		\For{$v \in \mathcal{V}$}
		\State $\mathbf{h}_v^{\!(0)\!}\! \gets\! \bm{\Theta}_v$, initialize the node-level representation of state $v$
		\EndFor
		\While{the parameters of \methodshort~ have not converged}
		\State take $N$ samples of $\left\{ \langle\mathbf{X}_{1:T}, \mathbf{Y}_{1:T}\rangle, \langle\mathbf{G}^{(1:T)}\rangle \right\}$ as a batch
		\For{each $\mathbf{X}_t \in \mathbf{X}_{1:T}$, $\mathbf{Y}_t \in \mathbf{Y}_{1:T}$, $\mathbf{G}^{(t)} \in \mathbf{G}^{(1:T)}$}
		\State $\left\{\mathbf{H}_v^{(t)}\right\}_{v\in\mathcal{V}} \gets \mathcal{F}_{\rm{MP}}\left(\mathbf{G}^{(t)}, \left\{\mathbf{h}_v^{(t-1)}\right\}_{v\in\mathcal{V}} \right)$, conduct message passing as \equationref{eq:edgerelation}
		\State $\alpha_t\gets$ compute attention score as \equationref{eq:hierarchical}.3
		\State $\left\{\mathbf{h}_v^{(t)}\right\}_{v\in\mathcal{V}} \gets$ node-level propagation as \equationref{eq:hierarchical}.1
		\State $\mathbf{U}^{(t)} \gets$ graph-level propagation as \equationref{eq:hierarchical}.2
		\State $\mathbf{h}_{\mathbf{G}}^{t} \gets$ compute current feature embedding as \equationref{eq:output} 
		\State $\mathbf{P}\left(\mathbf{Y}'_{t+1} | \mathbf{h}_{\mathbf{G}}^{t} \right) \gets $ estimate the probabilities of the next event
		\EndFor
		\State $\theta \gets \nabla_\theta \left[\frac{1}{N}\sum_{n=1}^{N} (\mathcal{L})\right]$, back-propagate the loss and train the whole \methodshort as \equationref{eq:loss}
		\EndWhile
		
	\end{algorithmic}
	\label{alg:0}
\end{algorithm}

\subsection{Implementation of State Recognition}
\label{sec:appendix:state}
In this section, 
we present several implementations for state recognition, including sequence clustering \citep{hallac2017toeplitz}, SAX words \citep{lin2007experiencing} and Shapelets \citep{lines2012shapelet}, which have been proven to be competitive for capturing the representative patterns (or \textit{states}), in previous works.

\vpara{Sequence Clustering.}
Cluster methods allow us to find the repeated patterns in time-series segments, which can reduce the dimension and allow us to derive insights capable of explaining time-series evolution \citep{hallac2017toeplitz, ailliot2012markov-switching, hu2019capturing}.
Herein, we take Kmeans\citep{kanungo2002an} as example; the aim here is to partition the $n$ segments $\mathbf{X}_t$ into $|\mathcal{V}|$ sets $\bm{\Omega} = \{\bm{\Omega}_1, ..., \bm{\Omega}_{|\mathcal{V}|}\}$, so as to minimize the within-cluster sum of squares, i.e., variance. Formally, the objective is to find:
\small
\begin{equation}
\mathop{\arg\min}_{\bm{\Omega} }  \sum_{v \in \mathcal{V}} \sum_{\mathbf{X}_t \in \bm{\Omega}_v } || \mathbf{X}_t - \bm{\Theta}_v ||^2 = \mathop{\arg\min}_{\bm{\Omega} } \sum_{v \in \mathcal{V}} |\bm{\Omega}_v| {\rm{Var}}~\bm{\Omega}_v
\end{equation}
\normalsize
where $\bm{\Theta}_v$ is the mean of all segments in $\bm{\Omega}_v$.
We then normalize the distance $\mathcal{D}(\mathbf{X}_t, \bm{\Theta}_v)$ between a segment $\mathbf{X}_t$ and patterns $\bm{\Theta}_v$ as the recognition weight, which can be formulated as follows:
\small
\begin{equation}
\begin{aligned}
&\mathcal{D}(\mathbf{X}_t, \bm{\Theta}_v) = || \mathbf{X}_t - \bm{\Theta}_v ||^2 \\
&\mathbf{P}(\bm{\Theta}_v | \mathbf{X}_t)  = \frac{\max([\mathcal{D}(\mathbf{X}_t, \bm{\Theta}_v)]_{v \in \mathcal{V}}) - \mathcal{D}(\mathbf{X}_t, \bm{\Theta}_v)} {\max([\mathcal{D}(\mathbf{X}_t, \bm{\Theta}_v)]_{v \in \mathcal{V}}) - \min([\mathcal{D}(\mathbf{X}_t, \bm{\Theta}_v)]_{v \in \mathcal{V}})}
\label{eq:state:distance}
\end{aligned}
\end{equation}
\normalsize
where we adopt Euclidean distance to measure the similarity between segment $\mathbf{X}_t$ and state patterns $\bm{\Theta}_v$; the smaller this distance, the more similar they are. We can then construct the \graphname to represent the relations among different clusters. 

\vpara{SAX word.}
Symbolic aggregate approximation (SAX) is the first symbolic representation for time series that allows for dimensional reduction and indexing with a lower-bounding distance measure.
It transforms the original time-series segments into several average values (PAA representation\footnote{https://jmotif.github.io/sax-vsm\_site/morea/algorithm/PAA.html}) and converts them into a string.

Herein, we can consider each SAX word as a state $v$ and extend the corresponding average value $a$ as representative patterns of the time series segments, i.e., $\bm{\Theta}_v = [a, ..., a]$.
Based on this, we can normalize the distance $\mathcal{D}(\mathbf{X}_t, \bm{\Theta}_v)$ as the recognition weight, following the approach outlined in \equationref{eq:state:define}.
Subsequently, we can construct the \graphname to represent the relations among SAX representations.

\vpara{Shapelet.}
A shapelet $\bm{\Theta}_v$ is a segment that is representative of a certain class. More precisely, it can separate segments into two smaller sets, one that is close to $\bm{\Theta}_v$ and another that is far from $\bm{\Theta}_v$ according to some specific criteria, such that for a given time series classification task, positive and negative samples can be put into different groups. The criteria for these can be formalized as
\begin{equation}
\mathcal{L}_{\rm{shapelet}} = -g\left(\mathcal{D}_{\rm{pos}}(\mathbf{X}_t, \bm{\Theta}_v), \mathcal{D}_{\rm{neg}}(\mathbf{X}_t, \bm{\Theta}_v)\right)
\end{equation}
where $\mathcal{L}_{\rm{shapelet}}$ measures the dissimilarity between positive and negative samples towards the shapelet $\bm{\Theta}_v$. $\mathcal{D}_{*}(\mathbf{X}_t, \bm{\Theta}_v)$ denotes the set of distances with respect to a specific group, i.e., positive or negative class; the function $g$ takes two finite sets as input and returns a scalar value to indicate how far apart these two sets are. This could be information gain or some dissimilarity measurements on sets (i.e., KL divergence). We can then adopt the same approaches as in the above definitions to recognize states' weights and construct the \graphname to represent the relations among shapelets. 

\subsection{Implementation of Message Passing}
\label{sec:appendix:gnn}
As for the implementations of message passing in local information aggregation, there are many existing works addressing this issue, such as pooling, GGNN \citep{li2016gated}, GCN \citep{duvenaud2015convolutional}, GraphSAGE \citep{hamilton2017inductive}, GAT \citep{gat2018petar}, etc.. Herein, we present their implementation details. 
Broadly speaking, the aim of message passing is to aggregate the \textit{messages} of node $v$'s neighbors, and thus to compute its new representation vector, the scheme of which is 
\begin{equation}
\mathbf{H}_v^{(t)} = \sum_{v' \in N(v)} \mathcal{F}_{\rm{MP}}\left(\mathbf{h}_{v'}^{(t-1)}, e_{(v, v')}^{(t)} \right)
\end{equation}
where $\mathbf{H}_v^{(t)}$ is the intermediate representation of node $v$ after aggregation; moreover, $\mathcal{F}_{\rm{MP}}(\cdot, \cdot)$ is the specific message function, which combines the messages from all $v$'s neighbors $N(v)$ in graph $\mathbf{G}^{(t)}$.

\vpara{Pooling.}
Pooling is a simple implementation, which receives the neighbors' messages by computing the production of these neighbors' representation and current transition weight. This approach can be formulated as
\begin{equation}
\mathcal{F}_{\rm{MP}}\left(\mathbf{h}_{v'}^{(t-1)}, e_{(v, v')}^{(t)} \right)= m_{(v, v')}^{(t)} \times\mathbf{h}_{v'}^{(t-1)}
\end{equation}
where $v' \in N(v)$ is a neighbor of node $v$ and $\mathbf{h}_{v'}^{(t-1)}$ is its representation of the last temporal point. $m_{(v, v')}^{(t)}$ is the current relation weight, which is computed by \equationref{eq:graph:build} (see details in \secref{sec:setup:graph}).

\vpara{GGNN.}
Gated Graph Neural Networks \citep{li2016gated} implement a message-feedback mechanism: in short, when node $v'$ passes a message to node $v$ via edge $(v' \rightarrow v)$, $v$ will send a feedback message to $v'$. This approach aggregates the in-degree and out-degree messages from its neighbors, which is formulated as
\begin{equation}
\begin{split}
\mathcal{F}_{\rm{MP}}\left(\mathbf{h}_{v'}^{(t-1)}, e_{(v, v')}^{(t)} \right)= W_{\rm{in}} \cdot\left[m_{(v', v)}^{(t)} \times\mathbf{h}_{v'}^{(t-1)} \right] +& \\
W_{\rm{out}} \cdot\left[m_{(v, v')}^{(t)} \times\mathbf{h}_{v}^{(t-1)} \right] +& ~b
\end{split}
\label{eq:appendix:ggnn}
\end{equation}
where $W, b$ is the learnable weight and bias, which is related to the downstream task. From the perspective of the whole graph (adjacency matrix), we in fact build a new graph with the opposite directed edges. Hence, the above scheme can be reformulated as
\begin{subequations}
	\begin{align}
	\mathcal{M}^{(t)} &= \left[ \begin{aligned}& \mathcal{M}_{in}^{(t)} \\ & \mathcal{M}_{out}^{(t)}\end{aligned} \right] = \left[ \begin{aligned}& \left[m_{(v, v')}^{(t)}\right]_{v, v' \in \mathcal{V}} \\ & \left[m_{(v, v')}^{(t)}\right]_{v, v' \in \mathcal{V}}^\top \end{aligned} \right]\\
	\begin{split}
	\mathbf{H}^{(t)} &=
	W \cdot \mathcal{M}^{(t)} \cdot \mathbf{h}^{(t-1)} + b \\
	&= [W_{in} ~W_{out}] \cdot \left[\begin{aligned}& \mathcal{M}_{in}^{(t)} \\ & \mathcal{M}_{out}^{(t)}\end{aligned}\right] \cdot \mathbf{h}^{(t-1)} + [b_{in} ~b_{out}] 
	\end{split}
	\end{align}
\end{subequations}
where $\mathcal{M}_{\rm{in}} = \left[m_{(v, v')}^{(t)}\right]_{v, v' \in \mathcal{V}} $ is the adjacency matrix in graph $\mathbf{G}^{(t)}$; ``$\top$'' indicates the transposition operator, i.e., $\mathcal{M}_{\rm{out}}$ is actually the transposition matrix of $\mathcal{M}_{\rm{in}}$.

\vpara{GCN.}
Graph Convolution Networks \citep{duvenaud2015convolutional} adopt spectral approaches to represent the graph. It computes the eigendecomposition of the graph Laplacian, defined as
\begin{equation}
\mathbf{H}^{(t)} = \bm{\mho}^{(t)} g(\bm{\Lambda}^{(t)}) {\bm{\mho}^{(t)}}^{\top} \cdot \mathbf{h}^{(t-1)}
\end{equation}
where $\bm{\mho}^{(t)}$ is the matrix of eigenvectors of the normalized graph Laplacian $\mathbf{L}^{(t)} = \mathbf{I}_{\mathcal{V}} - {\mathbf{D}^{(t)}}^{-\frac{1}{2}} \mathcal{M}^{(t)} {\mathbf{D}^{(t)}}^{-\frac{1}{2}} = \bm{\mho}^{(t)} g(\bm{\Lambda}^{(t)}) {\bm{\mho}^{(t)}}^{\top}$ ($\mathbf{D}^{(t)}$ is the degree matrix and $\mathcal{M}^{(t)}$ is the adjacency matrix of the graph $\mathbf{G}^{(t)}$), with a diagonal matrix of its eigenvalues $\bm{\Lambda}^{(t)}$. $g(\cdot)$ is the filter function, which can be approximated by a truncated expansion in terms of Chebyshev polynomials \citep{defferrard2016convolutional}.

\vpara{GraphSAGE.} 
In order to avoid transductive learning and naturally generalize to unseen nodes, \citet{hamilton2017inductive} proposed the general inductive framework, GraphSAGE, which generates new representation by sampling and aggregating features from a node's local neighborhood. The difference between this approach and the aforementioned GGNN (\equationref{eq:appendix:ggnn}) is that the former does not utilize the full set of neighbors, but rather fixed-size set of neighbors through uniform sampling.

\vpara{GAT.}
Graph Attention Networks adopt a \textit{self-attention} strategy, which involves computing the representations of each node attending to it over its neighbors. The attention coefficients are computed in the node pair $(v, v')$
\small
\begin{equation}
\alpha^{(t)}_{(v, v')} = \frac{{\rm{exp}}\left({\rm{LeakyReLU}}\left(W\left(\mathbf{h}_v^{(t-1)} \oplus \mathbf{h}_{v'}^{(t-1)}\right)\right)\right)}{\sum_{v'' \in N(v)}{\rm{exp}}\left({\rm{LeakyReLU}}\left(W\left(\mathbf{h}_v^{(t-1)} \oplus \mathbf{h}_{v''}^{(t-1)}\right)\right)\right)}
\end{equation}
\normalsize
where $\alpha^{(t)}_{(v, v')} $ is the attention coefficient of node $v$ and $v'$ in $\mathbf{G}^{(t)}$, which reweights the edge $m_{(v, v')}^{(t)}$. We can then adopt an approach similar to \equationref{eq:aggregation} to obtain $\mathbf{H}_v^{(t)}$ of each node.

\subsection{Hyperparameter Settings}
\label{sec:appendix:setup}
We have discussed several important hyperparameter settings of the proposed model in \secref{sec:exp:pa}. 
We conduct \textit{grid search} for our proposed model and baselines in order to find the adaptive hyperparameters and compare fairly.
The remaining aspects of parameter options are introduced below to facilitate better reproductivity.

\vpara{Hyperparameters in \methodshort. }
We test \methodshort at the number of states $|\mathcal{V}|$, segment length $\tau$, the size of graph-level representation $|\mathbf{U}|$ (the size of node-level representation $|\mathbf{h}|$ is determined by state recognition, since $\mathbf{h}_v^{(0)}=\bm{\Theta_v}$), while the search space may differ between different datasets. We test $|\mathcal{V}|$ with values from 5 to 100 with interval 10, and further test $\tau$ with different lengths that are smaller or greater than the period length of the corresponding dataset. We test $|\mathbf{U}|$ from $2^4$ to $2^{10}$ with exponential interval 1. In batch-wise training for \methodshort, the batch size is set to 1000, and we choose the Adam algorithm \citep{kingma2015adam} as the loss optimizer.

\vpara{Hyperparameters in baselines. }
As for baselines, we use the source code provided on \textit{TSLearn}\footnote{https://tslearn.readthedocs.io/en/latest} for several feature-based models, and code the sequential models by ourselves. For the graphical models, we conduct the experiments on the provided codes in GitHub.
If the parameter interface is open, we adopt the same grid search approach to search the best parameters.
Due to the binary event prediction tasks, we use XGBoost \citep{chen2016xgboost} with same parameters for all methods in order to improve the overall performance.

%% file: evolutiongraph.bbl

\begin{thebibliography}{46}


\ifx \showCODEN    \undefined \def \showCODEN     #1{\unskip}     \fi
\ifx \showDOI      \undefined \def \showDOI       #1{#1}\fi
\ifx \showISBNx    \undefined \def \showISBNx     #1{\unskip}     \fi
\ifx \showISBNxiii \undefined \def \showISBNxiii  #1{\unskip}     \fi
\ifx \showISSN     \undefined \def \showISSN      #1{\unskip}     \fi
\ifx \showLCCN     \undefined \def \showLCCN      #1{\unskip}     \fi
\ifx \shownote     \undefined \def \shownote      #1{#1}          \fi
\ifx \showarticletitle \undefined \def \showarticletitle #1{#1}   \fi
\ifx \showURL      \undefined \def \showURL       {\relax}        \fi
\providecommand\bibfield[2]{#2}
\providecommand\bibinfo[2]{#2}
\providecommand\natexlab[1]{#1}
\providecommand\showeprint[2][]{arXiv:#2}

\bibitem[\protect\citeauthoryear{Ailliot and Monbet}{Ailliot and
  Monbet}{2012}]%
        {ailliot2012markov-switching}
\bibfield{author}{\bibinfo{person}{Pierre Ailliot} {and}
  \bibinfo{person}{Valerie Monbet}.} \bibinfo{year}{2012}\natexlab{}.
\newblock \showarticletitle{Markov-switching autoregressive models for wind
  time series}.
\newblock \bibinfo{journal}{\emph{Environmental Modelling and Software}}
  \bibinfo{volume}{30} (\bibinfo{year}{2012}), \bibinfo{pages}{92--101}.
\newblock


\bibitem[\protect\citeauthoryear{Bagnall and Janacek}{Bagnall and
  Janacek}{2014}]%
        {Bagnall2014A}
\bibfield{author}{\bibinfo{person}{Anthony Bagnall} {and}
  \bibinfo{person}{Gareth Janacek}.} \bibinfo{year}{2014}\natexlab{}.
\newblock \showarticletitle{A Run Length Transformation for Discriminating
  Between Auto Regressive Time Series}.
\newblock \bibinfo{journal}{\emph{Journal of Classification}}
  (\bibinfo{year}{2014}), \bibinfo{pages}{154--178}.
\newblock


\bibitem[\protect\citeauthoryear{Bagnall, Lines, Bostrom, Large, and
  Keogh}{Bagnall et~al\mbox{.}}{2017}]%
        {bagnall2017the}
\bibfield{author}{\bibinfo{person}{Anthony~J Bagnall}, \bibinfo{person}{Jason
  Lines}, \bibinfo{person}{Aaron Bostrom}, \bibinfo{person}{James Large}, {and}
  \bibinfo{person}{Eamonn~J Keogh}.} \bibinfo{year}{2017}\natexlab{}.
\newblock \showarticletitle{The great time series classification bake off: a
  review and experimental evaluation of recent algorithmic advances}.
\newblock \bibinfo{journal}{\emph{DMKD}} \bibinfo{volume}{31},
  \bibinfo{number}{3} (\bibinfo{year}{2017}), \bibinfo{pages}{606--660}.
\newblock


\bibitem[\protect\citeauthoryear{Battaglia, Hamrick, Bapst, Sanchezgonzalez,
  Zambaldi, Malinowski, Tacchetti, Raposo, Santoro, Faulkner,
  et~al\mbox{.}}{Battaglia et~al\mbox{.}}{2018}]%
        {battaglia2018relational}
\bibfield{author}{\bibinfo{person}{Peter Battaglia}, \bibinfo{person}{Jessica~B
  Hamrick}, \bibinfo{person}{Victor Bapst}, \bibinfo{person}{Alvaro
  Sanchezgonzalez}, \bibinfo{person}{Vinicius~Flores Zambaldi},
  \bibinfo{person}{Mateusz Malinowski}, \bibinfo{person}{Andrea Tacchetti},
  \bibinfo{person}{David Raposo}, \bibinfo{person}{Adam Santoro},
  \bibinfo{person}{Ryan Faulkner}, {et~al\mbox{.}}}
  \bibinfo{year}{2018}\natexlab{}.
\newblock \showarticletitle{Relational inductive biases, deep learning, and
  graph networks}.
\newblock \bibinfo{journal}{\emph{arXiv: Learning}} (\bibinfo{year}{2018}).
\newblock


\bibitem[\protect\citeauthoryear{Battaglia, Pascanu, Lai, Rezende, and
  Kavukcuoglu}{Battaglia et~al\mbox{.}}{2016}]%
        {battaglia2016interaction}
\bibfield{author}{\bibinfo{person}{Peter Battaglia}, \bibinfo{person}{Razvan
  Pascanu}, \bibinfo{person}{Matthew Lai}, \bibinfo{person}{Danilo~Jimenez
  Rezende}, {and} \bibinfo{person}{Koray Kavukcuoglu}.}
  \bibinfo{year}{2016}\natexlab{}.
\newblock \showarticletitle{Interaction networks for learning about objects,
  relations and physics}.
\newblock \bibinfo{journal}{\emph{NeurIPS}} (\bibinfo{year}{2016}),
  \bibinfo{pages}{4509--4517}.
\newblock


\bibitem[\protect\citeauthoryear{Bengio, Simard, and Frasconi}{Bengio
  et~al\mbox{.}}{1994}]%
        {bengio1994learning}
\bibfield{author}{\bibinfo{person}{Yoshua Bengio}, \bibinfo{person}{Patrice~Y
  Simard}, {and} \bibinfo{person}{Paolo Frasconi}.}
  \bibinfo{year}{1994}\natexlab{}.
\newblock \showarticletitle{Learning long-term dependencies with gradient
  descent is difficult}.
\newblock \bibinfo{journal}{\emph{TNNLS}} \bibinfo{volume}{5},
  \bibinfo{number}{2} (\bibinfo{year}{1994}), \bibinfo{pages}{157--166}.
\newblock


\bibitem[\protect\citeauthoryear{Bouttefroy, Bouzerdoum, Phung, and
  Beghdadi}{Bouttefroy et~al\mbox{.}}{2010}]%
        {bouttefroy2010on}
\bibfield{author}{\bibinfo{person}{Philippe Loic~Marie Bouttefroy},
  \bibinfo{person}{Abdesselam Bouzerdoum}, \bibinfo{person}{Son~Lam Phung},
  {and} \bibinfo{person}{Azeddine Beghdadi}.} \bibinfo{year}{2010}\natexlab{}.
\newblock \showarticletitle{On the analysis of background subtraction
  techniques using Gaussian Mixture Models}.
\newblock \bibinfo{journal}{\emph{ICASSP}} (\bibinfo{year}{2010}),
  \bibinfo{pages}{4042--4045}.
\newblock


\bibitem[\protect\citeauthoryear{Brandes}{Brandes}{2001}]%
        {brandes2001a}
\bibfield{author}{\bibinfo{person}{Ulrik Brandes}.}
  \bibinfo{year}{2001}\natexlab{}.
\newblock \showarticletitle{A Faster Algorithm for Betweenness Centrality}.
\newblock \bibinfo{journal}{\emph{Mathematical Sociology}}
  \bibinfo{volume}{25}, \bibinfo{number}{2} (\bibinfo{year}{2001}),
  \bibinfo{pages}{163--177}.
\newblock


\bibitem[\protect\citeauthoryear{Chen and Guestrin}{Chen and Guestrin}{2016}]%
        {chen2016xgboost}
\bibfield{author}{\bibinfo{person}{Tianqi Chen} {and} \bibinfo{person}{Carlos
  Guestrin}.} \bibinfo{year}{2016}\natexlab{}.
\newblock \showarticletitle{XGBoost: A Scalable Tree Boosting System}.
\newblock \bibinfo{journal}{\emph{SIGKDD}} (\bibinfo{year}{2016}),
  \bibinfo{pages}{785--794}.
\newblock


\bibitem[\protect\citeauthoryear{Cheng, Yang, Wang, Hu, Zhuang, and Song}{Cheng
  et~al\mbox{.}}{2020}]%
        {cheng2019time2graph}
\bibfield{author}{\bibinfo{person}{Ziqiang Cheng}, \bibinfo{person}{Yang Yang},
  \bibinfo{person}{Wei Wang}, \bibinfo{person}{Wenjie Hu},
  \bibinfo{person}{Yueting Zhuang}, {and} \bibinfo{person}{Guojie Song}.}
  \bibinfo{year}{2020}\natexlab{}.
\newblock \showarticletitle{Time2Graph: Revisiting Time Series Modeling with
  Dynamic Shapelets}.
\newblock \bibinfo{journal}{\emph{AAAI}} (\bibinfo{year}{2020}),
  \bibinfo{pages}{3617--3624}.
\newblock


\bibitem[\protect\citeauthoryear{Chung, Ahn, and Bengio}{Chung
  et~al\mbox{.}}{2017}]%
        {chung2017hierarchical}
\bibfield{author}{\bibinfo{person}{Junyoung Chung}, \bibinfo{person}{Sungjin
  Ahn}, {and} \bibinfo{person}{Yoshua Bengio}.}
  \bibinfo{year}{2017}\natexlab{}.
\newblock \showarticletitle{Hierarchical Multiscale Recurrent Neural Networks}.
\newblock \bibinfo{journal}{\emph{ICLR}} (\bibinfo{year}{2017}).
\newblock


\bibitem[\protect\citeauthoryear{Chung, Gulcehre, Cho, and Bengio}{Chung
  et~al\mbox{.}}{2015}]%
        {chung2015gated}
\bibfield{author}{\bibinfo{person}{Junyoung Chung}, \bibinfo{person}{Caglar
  Gulcehre}, \bibinfo{person}{Kyunghyun Cho}, {and} \bibinfo{person}{Yoshua
  Bengio}.} \bibinfo{year}{2015}\natexlab{}.
\newblock \showarticletitle{Gated Feedback Recurrent Neural Networks}.
\newblock \bibinfo{journal}{\emph{ICML}} (\bibinfo{year}{2015}),
  \bibinfo{pages}{2067--2075}.
\newblock


\bibitem[\protect\citeauthoryear{Defferrard, Bresson, and
  Vandergheynst}{Defferrard et~al\mbox{.}}{2016}]%
        {defferrard2016convolutional}
\bibfield{author}{\bibinfo{person}{Michael Defferrard}, \bibinfo{person}{Xavier
  Bresson}, {and} \bibinfo{person}{Pierre Vandergheynst}.}
  \bibinfo{year}{2016}\natexlab{}.
\newblock \showarticletitle{Convolutional neural networks on graphs with fast
  localized spectral filtering}.
\newblock \bibinfo{journal}{\emph{NeurIPS}} (\bibinfo{year}{2016}),
  \bibinfo{pages}{3844--3852}.
\newblock


\bibitem[\protect\citeauthoryear{Du, Dai, Trivedi, Upadhyay, Gomezrodriguez,
  and Song}{Du et~al\mbox{.}}{2016}]%
        {du2016recurrent}
\bibfield{author}{\bibinfo{person}{Nan Du}, \bibinfo{person}{Hanjun Dai},
  \bibinfo{person}{Rakshit Trivedi}, \bibinfo{person}{Utkarsh Upadhyay},
  \bibinfo{person}{Manuel Gomezrodriguez}, {and} \bibinfo{person}{Le Song}.}
  \bibinfo{year}{2016}\natexlab{}.
\newblock \showarticletitle{Recurrent Marked Temporal Point Processes:
  Embedding Event History to Vector}.
\newblock \bibinfo{journal}{\emph{SIGKDD}} (\bibinfo{year}{2016}),
  \bibinfo{pages}{1555--1564}.
\newblock


\bibitem[\protect\citeauthoryear{Duvenaud, Maclaurin, Iparraguirre, Bombarell,
  Hirzel, Aspuru-Guzik, and Adams}{Duvenaud et~al\mbox{.}}{2015}]%
        {duvenaud2015convolutional}
\bibfield{author}{\bibinfo{person}{David~K Duvenaud}, \bibinfo{person}{Dougal
  Maclaurin}, \bibinfo{person}{Jorge Iparraguirre}, \bibinfo{person}{Rafael
  Bombarell}, \bibinfo{person}{Timothy Hirzel}, \bibinfo{person}{Al{\'a}n
  Aspuru-Guzik}, {and} \bibinfo{person}{Ryan~P Adams}.}
  \bibinfo{year}{2015}\natexlab{}.
\newblock \showarticletitle{Convolutional networks on graphs for learning
  molecular fingerprints}. In \bibinfo{booktitle}{\emph{NeurIPS}}.
  \bibinfo{pages}{2224--2232}.
\newblock


\bibitem[\protect\citeauthoryear{Feng, Yang, Hu, Wu, and Zhuang}{Feng
  et~al\mbox{.}}{2018}]%
        {feng2018scalefree}
\bibfield{author}{\bibinfo{person}{Rui Feng}, \bibinfo{person}{Yang Yang},
  \bibinfo{person}{Wenjie Hu}, \bibinfo{person}{Fei Wu}, {and}
  \bibinfo{person}{Yueting Zhuang}.} \bibinfo{year}{2018}\natexlab{}.
\newblock \showarticletitle{Representation Learning for Scale-free Networks}.
\newblock \bibinfo{journal}{\emph{AAAI}} (\bibinfo{year}{2018}),
  \bibinfo{pages}{282--289}.
\newblock


\bibitem[\protect\citeauthoryear{Fout, Byrd, Shariat, and Ben-Hur}{Fout
  et~al\mbox{.}}{2017}]%
        {fout2017protein}
\bibfield{author}{\bibinfo{person}{Alex Fout}, \bibinfo{person}{Jonathon Byrd},
  \bibinfo{person}{Basir Shariat}, {and} \bibinfo{person}{Asa Ben-Hur}.}
  \bibinfo{year}{2017}\natexlab{}.
\newblock \showarticletitle{Protein interface prediction using graph
  convolutional networks}. In \bibinfo{booktitle}{\emph{NeurIPS}}.
  \bibinfo{pages}{6530--6539}.
\newblock


\bibitem[\protect\citeauthoryear{Geng, Li, Wang, Zhang, Ye, Liu, and Yang}{Geng
  et~al\mbox{.}}{2019}]%
        {geng2019spatiotemporal}
\bibfield{author}{\bibinfo{person}{Xu Geng}, \bibinfo{person}{Yaguang Li},
  \bibinfo{person}{Leye Wang}, \bibinfo{person}{Lingyu Zhang},
  \bibinfo{person}{Jieping Ye}, \bibinfo{person}{Yan Liu}, {and}
  \bibinfo{person}{Qiang Yang}.} \bibinfo{year}{2019}\natexlab{}.
\newblock \showarticletitle{Spatiotemporal Multi-Graph Convolution Network for
  Ride-hailing Demand Forecasting}.
\newblock \bibinfo{journal}{\emph{AAAI}}  \bibinfo{volume}{33}
  (\bibinfo{year}{2019}), \bibinfo{pages}{3656--3663}.
\newblock


\bibitem[\protect\citeauthoryear{Gilmer, Schoenholz, Riley, Vinyals, and
  Dahl}{Gilmer et~al\mbox{.}}{2017}]%
        {gilmer2017neural}
\bibfield{author}{\bibinfo{person}{Justin Gilmer}, \bibinfo{person}{Samuel~S
  Schoenholz}, \bibinfo{person}{Patrick~F Riley}, \bibinfo{person}{Oriol
  Vinyals}, {and} \bibinfo{person}{George~E Dahl}.}
  \bibinfo{year}{2017}\natexlab{}.
\newblock \showarticletitle{Neural Message Passing for Quantum Chemistry}.
\newblock \bibinfo{journal}{\emph{ICML}} (\bibinfo{year}{2017}),
  \bibinfo{pages}{1263--1272}.
\newblock


\bibitem[\protect\citeauthoryear{Hallac, Vare, Boyd, and Leskovec}{Hallac
  et~al\mbox{.}}{2017}]%
        {hallac2017toeplitz}
\bibfield{author}{\bibinfo{person}{David Hallac}, \bibinfo{person}{Sagar Vare},
  \bibinfo{person}{Stephen~P Boyd}, {and} \bibinfo{person}{Jure Leskovec}.}
  \bibinfo{year}{2017}\natexlab{}.
\newblock \showarticletitle{Toeplitz Inverse Covariance-Based Clustering of
  Multivariate Time Series Data}.
\newblock \bibinfo{journal}{\emph{SIGKDD}} (\bibinfo{year}{2017}),
  \bibinfo{pages}{215--223}.
\newblock


\bibitem[\protect\citeauthoryear{Hamaguchi, Oiwa, Shimbo, and
  Matsumoto}{Hamaguchi et~al\mbox{.}}{2017}]%
        {hamaguchi2017knowledge}
\bibfield{author}{\bibinfo{person}{Takuo Hamaguchi}, \bibinfo{person}{Hidekazu
  Oiwa}, \bibinfo{person}{Masashi Shimbo}, {and} \bibinfo{person}{Yuji
  Matsumoto}.} \bibinfo{year}{2017}\natexlab{}.
\newblock \showarticletitle{Knowledge Transfer for Out-of-Knowledge-Base
  Entities : A Graph Neural Network Approach}.
\newblock \bibinfo{journal}{\emph{IJCAI}} (\bibinfo{year}{2017}),
  \bibinfo{pages}{1802--1808}.
\newblock


\bibitem[\protect\citeauthoryear{Hamilton, Ying, and Leskovec}{Hamilton
  et~al\mbox{.}}{2017}]%
        {hamilton2017inductive}
\bibfield{author}{\bibinfo{person}{William~L Hamilton}, \bibinfo{person}{Rex
  Ying}, {and} \bibinfo{person}{Jure Leskovec}.}
  \bibinfo{year}{2017}\natexlab{}.
\newblock \showarticletitle{Inductive Representation Learning on Large Graphs}.
\newblock \bibinfo{journal}{\emph{NeurIPS}} (\bibinfo{year}{2017}).
\newblock


\bibitem[\protect\citeauthoryear{Hochreiter and Schmidhuber}{Hochreiter and
  Schmidhuber}{1997}]%
        {hochreiter1997long}
\bibfield{author}{\bibinfo{person}{Sepp Hochreiter} {and}
  \bibinfo{person}{Jurgen Schmidhuber}.} \bibinfo{year}{1997}\natexlab{}.
\newblock \showarticletitle{Long Short-Term Memory}.
\newblock \bibinfo{journal}{\emph{Neural Computation}} (\bibinfo{year}{1997}),
  \bibinfo{pages}{1735--1780}.
\newblock


\bibitem[\protect\citeauthoryear{Hu, Yang, Wang, Huang, and Cheng}{Hu
  et~al\mbox{.}}{2020}]%
        {hu2020theft}
\bibfield{author}{\bibinfo{person}{Wenjie Hu}, \bibinfo{person}{Yang Yang},
  \bibinfo{person}{Jianbo Wang}, \bibinfo{person}{Xuanwen Huang}, {and}
  \bibinfo{person}{Ziqiang Cheng}.} \bibinfo{year}{2020}\natexlab{}.
\newblock \showarticletitle{Understanding Electricity-Theft Behavior via
  Multi-Source Data}.
\newblock \bibinfo{journal}{\emph{WWW}} (\bibinfo{year}{2020}),
  \bibinfo{pages}{2264--2274}.
\newblock


\bibitem[\protect\citeauthoryear{Kanungo, Mount, Netanyahu, Piatko, Silverman,
  and Wu}{Kanungo et~al\mbox{.}}{2002}]%
        {kanungo2002an}
\bibfield{author}{\bibinfo{person}{Tapas Kanungo}, \bibinfo{person}{David~M
  Mount}, \bibinfo{person}{Nathan~S Netanyahu}, \bibinfo{person}{Christine~D
  Piatko}, \bibinfo{person}{Ruth Silverman}, {and} \bibinfo{person}{Angela~Y
  Wu}.} \bibinfo{year}{2002}\natexlab{}.
\newblock \showarticletitle{An efficient k-means clustering algorithm: analysis
  and implementation}.
\newblock \bibinfo{journal}{\emph{TPAMI}} \bibinfo{volume}{24},
  \bibinfo{number}{7} (\bibinfo{year}{2002}), \bibinfo{pages}{881--892}.
\newblock


\bibitem[\protect\citeauthoryear{Kingma and Ba}{Kingma and Ba}{2015}]%
        {kingma2015adam}
\bibfield{author}{\bibinfo{person}{Diederik~P Kingma} {and}
  \bibinfo{person}{Jimmy Ba}.} \bibinfo{year}{2015}\natexlab{}.
\newblock \showarticletitle{Adam: A Method for Stochastic Optimization}.
\newblock \bibinfo{journal}{\emph{ICLR}} (\bibinfo{year}{2015}).
\newblock


\bibitem[\protect\citeauthoryear{Li, Tarlow, Brockschmidt, and Zemel}{Li
  et~al\mbox{.}}{2016}]%
        {li2016gated}
\bibfield{author}{\bibinfo{person}{Yujia Li}, \bibinfo{person}{Daniel Tarlow},
  \bibinfo{person}{Marc Brockschmidt}, {and} \bibinfo{person}{Richard~S
  Zemel}.} \bibinfo{year}{2016}\natexlab{}.
\newblock \showarticletitle{Gated Graph Sequence Neural Networks}.
\newblock \bibinfo{journal}{\emph{ICLR}} (\bibinfo{year}{2016}).
\newblock


\bibitem[\protect\citeauthoryear{Lin, Keogh, Wei, and Lonardi}{Lin
  et~al\mbox{.}}{2007}]%
        {lin2007experiencing}
\bibfield{author}{\bibinfo{person}{Jessica Lin}, \bibinfo{person}{Eamonn~J
  Keogh}, \bibinfo{person}{Li Wei}, {and} \bibinfo{person}{Stefano Lonardi}.}
  \bibinfo{year}{2007}\natexlab{}.
\newblock \showarticletitle{Experiencing SAX: a novel symbolic representation
  of time series}.
\newblock \bibinfo{journal}{\emph{DMKD}} \bibinfo{volume}{15},
  \bibinfo{number}{2} (\bibinfo{year}{2007}), \bibinfo{pages}{107--144}.
\newblock


\bibitem[\protect\citeauthoryear{Lin, Khade, and Li}{Lin et~al\mbox{.}}{2012}]%
        {Lin2012Rotation}
\bibfield{author}{\bibinfo{person}{Jessica Lin}, \bibinfo{person}{Rohan Khade},
  {and} \bibinfo{person}{Yuan Li}.} \bibinfo{year}{2012}\natexlab{}.
\newblock \showarticletitle{Rotation-invariant similarity in time series using
  bag-of-patterns representation}.
\newblock \bibinfo{journal}{\emph{IJIIS}} (\bibinfo{year}{2012}),
  \bibinfo{pages}{287--315}.
\newblock


\bibitem[\protect\citeauthoryear{Lines, Davis, Hills, and Bagnall}{Lines
  et~al\mbox{.}}{2012}]%
        {lines2012shapelet}
\bibfield{author}{\bibinfo{person}{Jason Lines}, \bibinfo{person}{Luke~M
  Davis}, \bibinfo{person}{Jon Hills}, {and} \bibinfo{person}{Anthony
  Bagnall}.} \bibinfo{year}{2012}\natexlab{}.
\newblock \showarticletitle{A shapelet transform for time series
  classification}. In \bibinfo{booktitle}{\emph{SIGKDD}}. ACM,
  \bibinfo{pages}{289--297}.
\newblock


\bibitem[\protect\citeauthoryear{Liu, Shi, Pierce, and Ren}{Liu
  et~al\mbox{.}}{2019}]%
        {liu2019characterizing}
\bibfield{author}{\bibinfo{person}{Yozen Liu}, \bibinfo{person}{Xiaolin Shi},
  \bibinfo{person}{Lucas Pierce}, {and} \bibinfo{person}{Xiang Ren}.}
  \bibinfo{year}{2019}\natexlab{}.
\newblock \showarticletitle{Characterizing and Forecasting User Engagement with
  In-app Action Graph: A Case Study of Snapchat}.
\newblock \bibinfo{journal}{\emph{SIGKDD}} (\bibinfo{year}{2019}),
  \bibinfo{pages}{2023--2031}.
\newblock


\bibitem[\protect\citeauthoryear{Ning, Muthiah, Rangwala, and
  Ramakrishnan}{Ning et~al\mbox{.}}{2016}]%
        {ning2016modeling}
\bibfield{author}{\bibinfo{person}{Yue Ning}, \bibinfo{person}{Sathappan
  Muthiah}, \bibinfo{person}{Huzefa Rangwala}, {and} \bibinfo{person}{Naren
  Ramakrishnan}.} \bibinfo{year}{2016}\natexlab{}.
\newblock \showarticletitle{Modeling Precursors for Event Forecasting via
  Nested Multi-Instance Learning}.
\newblock \bibinfo{journal}{\emph{SIGKDD}} (\bibinfo{year}{2016}),
  \bibinfo{pages}{1095--1104}.
\newblock


\bibitem[\protect\citeauthoryear{Opsahl, Agneessens, and Skvoretz}{Opsahl
  et~al\mbox{.}}{2010}]%
        {opsahl2010node}
\bibfield{author}{\bibinfo{person}{Tore Opsahl}, \bibinfo{person}{Filip
  Agneessens}, {and} \bibinfo{person}{John Skvoretz}.}
  \bibinfo{year}{2010}\natexlab{}.
\newblock \showarticletitle{Node centrality in weighted networks: Generalizing
  degree and shortest paths}.
\newblock \bibinfo{journal}{\emph{Social Networks}} \bibinfo{volume}{32},
  \bibinfo{number}{3} (\bibinfo{year}{2010}), \bibinfo{pages}{245--251}.
\newblock


\bibitem[\protect\citeauthoryear{Page, Brin, Motwani, and Winograd}{Page
  et~al\mbox{.}}{1999}]%
        {page1999the}
\bibfield{author}{\bibinfo{person}{Lawrence Page}, \bibinfo{person}{Sergey
  Brin}, \bibinfo{person}{Rajeev Motwani}, {and} \bibinfo{person}{Terry
  Winograd}.} \bibinfo{year}{1999}\natexlab{}.
\newblock \showarticletitle{The PageRank Citation Ranking: Bringing Order to
  the Web.}
\newblock \bibinfo{journal}{\emph{WWW}} (\bibinfo{year}{1999}),
  \bibinfo{pages}{161--172}.
\newblock


\bibitem[\protect\citeauthoryear{Pareja, Domeniconi, Chen, Ma, Suzumura,
  Kanezashi, Kaler, and Leisersen}{Pareja et~al\mbox{.}}{2020}]%
        {pareja2019evolvegcn}
\bibfield{author}{\bibinfo{person}{Aldo Pareja}, \bibinfo{person}{Giacomo
  Domeniconi}, \bibinfo{person}{Jie Chen}, \bibinfo{person}{Tengfei Ma},
  \bibinfo{person}{Toyotaro Suzumura}, \bibinfo{person}{Hiroki Kanezashi},
  \bibinfo{person}{Tim Kaler}, {and} \bibinfo{person}{Charles~E Leisersen}.}
  \bibinfo{year}{2020}\natexlab{}.
\newblock \showarticletitle{EvolveGCN: Evolving Graph Convolutional Networks
  for Dynamic Graphs.}
\newblock \bibinfo{journal}{\emph{AAAI}} (\bibinfo{year}{2020}).
\newblock


\bibitem[\protect\citeauthoryear{Pascanu, Mikolov, and Bengio}{Pascanu
  et~al\mbox{.}}{2013}]%
        {pascanu2013on}
\bibfield{author}{\bibinfo{person}{Razvan Pascanu}, \bibinfo{person}{Tomas
  Mikolov}, {and} \bibinfo{person}{Yoshua Bengio}.}
  \bibinfo{year}{2013}\natexlab{}.
\newblock \showarticletitle{On the difficulty of training recurrent neural
  networks}.
\newblock \bibinfo{journal}{\emph{ICML}} (\bibinfo{year}{2013}),
  \bibinfo{pages}{1310--1318}.
\newblock


\bibitem[\protect\citeauthoryear{Perozzi, Alrfou, and Skiena}{Perozzi
  et~al\mbox{.}}{2014}]%
        {perozzi2014deepwalk}
\bibfield{author}{\bibinfo{person}{Bryan Perozzi}, \bibinfo{person}{Rami
  Alrfou}, {and} \bibinfo{person}{Steven Skiena}.}
  \bibinfo{year}{2014}\natexlab{}.
\newblock \showarticletitle{DeepWalk: online learning of social
  representations}.
\newblock \bibinfo{journal}{\emph{SIGKDD}} (\bibinfo{year}{2014}),
  \bibinfo{pages}{701--710}.
\newblock


\bibitem[\protect\citeauthoryear{Rabiner and Juang}{Rabiner and Juang}{1986}]%
        {rabiner1986an}
\bibfield{author}{\bibinfo{person}{L~R Rabiner} {and}
  \bibinfo{person}{Biinghwang Juang}.} \bibinfo{year}{1986}\natexlab{}.
\newblock \showarticletitle{An introduction to hidden Markov models}.
\newblock \bibinfo{journal}{\emph{IEEE Assp Magazine}} \bibinfo{volume}{3},
  \bibinfo{number}{1} (\bibinfo{year}{1986}), \bibinfo{pages}{4--16}.
\newblock


\bibitem[\protect\citeauthoryear{Rakthanmanon and Keogh}{Rakthanmanon and
  Keogh}{2013}]%
        {rakthanmanon2013fast}
\bibfield{author}{\bibinfo{person}{Thanawin Rakthanmanon} {and}
  \bibinfo{person}{Eamonn Keogh}.} \bibinfo{year}{2013}\natexlab{}.
\newblock \showarticletitle{Fast shapelets: A scalable algorithm for
  discovering time series shapelets}.
\newblock \bibinfo{journal}{\emph{ICDM}} (\bibinfo{year}{2013}),
  \bibinfo{pages}{668--676}.
\newblock


\bibitem[\protect\citeauthoryear{Sanchez, Heess, Springenberg, Merel, Hadsell,
  Riedmiller, and Battaglia}{Sanchez et~al\mbox{.}}{2018}]%
        {sanchez2018graph}
\bibfield{author}{\bibinfo{person}{Alvaro Sanchez}, \bibinfo{person}{Nicolas
  Heess}, \bibinfo{person}{Jost~Tobias Springenberg}, \bibinfo{person}{Josh
  Merel}, \bibinfo{person}{Raia Hadsell}, \bibinfo{person}{Martin~A
  Riedmiller}, {and} \bibinfo{person}{Peter Battaglia}.}
  \bibinfo{year}{2018}\natexlab{}.
\newblock \showarticletitle{Graph Networks as Learnable Physics Engines for
  Inference and Control}.
\newblock \bibinfo{journal}{\emph{ICML}} (\bibinfo{year}{2018}),
  \bibinfo{pages}{4467--4476}.
\newblock


\bibitem[\protect\citeauthoryear{Senin and Malinchik}{Senin and
  Malinchik}{2013}]%
        {senin2013sax-vsm}
\bibfield{author}{\bibinfo{person}{Pavel Senin} {and} \bibinfo{person}{Sergey
  Malinchik}.} \bibinfo{year}{2013}\natexlab{}.
\newblock \showarticletitle{SAX-VSM: Interpretable Time Series Classification
  Using SAX and Vector Space Model}.
\newblock \bibinfo{journal}{\emph{ICDM}} (\bibinfo{year}{2013}),
  \bibinfo{pages}{1175--1180}.
\newblock


\bibitem[\protect\citeauthoryear{Velickovic, Cucurull, Casanova, Romero, Lio,
  and Bengio}{Velickovic et~al\mbox{.}}{2018}]%
        {gat2018petar}
\bibfield{author}{\bibinfo{person}{Petar Velickovic}, \bibinfo{person}{Guillem
  Cucurull}, \bibinfo{person}{Arantxa Casanova}, \bibinfo{person}{Adriana
  Romero}, \bibinfo{person}{Pietro Lio}, {and} \bibinfo{person}{Yoshua
  Bengio}.} \bibinfo{year}{2018}\natexlab{}.
\newblock \showarticletitle{Graph Attention Networks}.
\newblock \bibinfo{journal}{\emph{ICLR}} (\bibinfo{year}{2018}).
\newblock


\bibitem[\protect\citeauthoryear{Wang, Wang, Li, and Wu}{Wang
  et~al\mbox{.}}{2018b}]%
        {wang2018multilevel}
\bibfield{author}{\bibinfo{person}{Jingyuan Wang}, \bibinfo{person}{Ze Wang},
  \bibinfo{person}{Jianfeng Li}, {and} \bibinfo{person}{Junjie Wu}.}
  \bibinfo{year}{2018}\natexlab{b}.
\newblock \showarticletitle{Multilevel Wavelet Decomposition Network for
  Interpretable Time Series Analysis}.
\newblock \bibinfo{journal}{\emph{SIGKDD}} (\bibinfo{year}{2018}),
  \bibinfo{pages}{2437--2446}.
\newblock


\bibitem[\protect\citeauthoryear{Wang, Girshick, Gupta, and He}{Wang
  et~al\mbox{.}}{2018a}]%
        {wang2018non-local}
\bibfield{author}{\bibinfo{person}{Xiaolong Wang}, \bibinfo{person}{Ross~B
  Girshick}, \bibinfo{person}{Abhinav Gupta}, {and} \bibinfo{person}{Kaiming
  He}.} \bibinfo{year}{2018}\natexlab{a}.
\newblock \showarticletitle{Non-Local Neural Networks}.
\newblock \bibinfo{journal}{\emph{CVPR}} (\bibinfo{year}{2018}).
\newblock


\bibitem[\protect\citeauthoryear{Yang and Jiang}{Yang and Jiang}{2014}]%
        {Yang2014HMM}
\bibfield{author}{\bibinfo{person}{Yun Yang} {and} \bibinfo{person}{Jianmin
  Jiang}.} \bibinfo{year}{2014}\natexlab{}.
\newblock \showarticletitle{HMM-based hybrid meta-clustering ensemble for
  temporal data}.
\newblock \bibinfo{journal}{\emph{KBS}} (\bibinfo{year}{2014}),
  \bibinfo{pages}{299--310}.
\newblock


\bibitem[\protect\citeauthoryear{Zhou, Yang, Ren, Wu, and Zhuang}{Zhou
  et~al\mbox{.}}{2018}]%
        {zhou2018dynamic}
\bibfield{author}{\bibinfo{person}{Lekui Zhou}, \bibinfo{person}{Yang Yang},
  \bibinfo{person}{Xiang Ren}, \bibinfo{person}{Fei Wu}, {and}
  \bibinfo{person}{Yueting Zhuang}.} \bibinfo{year}{2018}\natexlab{}.
\newblock \showarticletitle{Dynamic Network Embedding by Modeling Triadic
  Closure Process}.
\newblock \bibinfo{journal}{\emph{AAAI}} (\bibinfo{year}{2018}),
  \bibinfo{pages}{571--578}.
\newblock


\end{thebibliography}
